\pdfoutput=1

\documentclass[11pt]{article}

\usepackage{acl}

\usepackage{times}
\usepackage{latexsym}

\usepackage{caption}
\usepackage{graphicx}
\usepackage{subcaption}
\usepackage{amsmath}
\usepackage{amsfonts}
\usepackage{booktabs}
\usepackage{algorithm}
\usepackage{algpseudocode}
\makeatletter
\newlength{\trianglerightwidth}
\algnewcommand{\LineComment}[1]{\Statex \hskip\ALG@thistlm
  \parbox[t]{\dimexpr\linewidth-\ALG@thistlm}{\hangindent=\algorithmicindent \hangafter=1 \strut\makebox[\algorithmicindent][l]{$\triangleright$}#1\strut}}
\algnewcommand{\LineCommentAfter}[1]{\Statex \hskip\ALG@tlm
  \parbox[t]{\dimexpr\linewidth-\ALG@tlm}{\hangindent=\algorithmicindent \hangafter=1 \strut\makebox[\algorithmicindent][l]{$\triangleright$}#1\strut}}
\makeatother

\usepackage{multirow}
\usepackage{multicol}
\usepackage{multibib}
\usepackage{pdfpages,comment}

\definecolor{darkgreen}{RGB}{0, 128, 0}

\usepackage[T1]{fontenc}

\usepackage[utf8]{inputenc}

\usepackage{microtype}
\title{An Embarrassingly Simple Approach for Intellectual Property Rights Protection on Recurrent Neural Networks}

\author{Zhi Qin Tan \and Hao Shan Wong \and Chee Seng Chan \\
CISiP, Universiti Malaya, Malaysia \\
\texttt{zhiqin1998@hotmail.com; haoshanw@gmail.com; cs.chan@um.edu.my }}

\begin{document}
\maketitle
\begin{abstract}
Capitalise on deep learning models, offering Natural Language Processing (NLP) solutions as a part of the Machine Learning as a Service (MLaaS) has generated handsome revenues. At the same time, it is known that the creation of these lucrative deep models is non-trivial. Therefore, protecting these inventions' intellectual property rights (IPR) from being abused, stolen and plagiarized is vital. This paper proposes a practical approach for the IPR protection on recurrent neural networks (RNN) without all the bells and whistles of existing IPR solutions. Particularly, we introduce the \emph{Gatekeeper} concept that resembles the recurrent nature in RNN architecture to embed keys. Also, we design the model training scheme in a way such that the protected RNN model will retain its original performance \emph{iff} a genuine key is presented. Extensive experiments showed that our protection scheme is \emph{robust} and \emph{effective} against ambiguity and removal attacks in both white-box and black-box protection schemes on different RNN variants. Code is available at \url{https://github.com/zhiqin1998/RecurrentIPR}.

\end{abstract}

\section{Introduction}

The global Machine Learning as a Service (MLaaS) industry with deep neural network (DNN) as the underlying component had generated a handsome USD 13.95 billion revenue in 2020 and is expected to reach USD 302.66 billion by 2030, witnessing a Compound Annual Growth Rate (CAGR)\footnote{The mean annual growth rate of an investment over a specified period of time longer than one year.} of 36.2\% from 2021 to 2030 \cite{mlmarket2022}. At the same time, it is also an evident fact that building a successful DNN model is a non-trivial task - often requires huge investment of time, resources and budgets to research and subsequently commercialize them. As such, the creation of such DNN models should be well protected to prevent them from being replicated, redistributed or shared by illegal parties.

\begin{figure}[t]
\centering
\includegraphics[keepaspectratio=true, scale = 0.25]{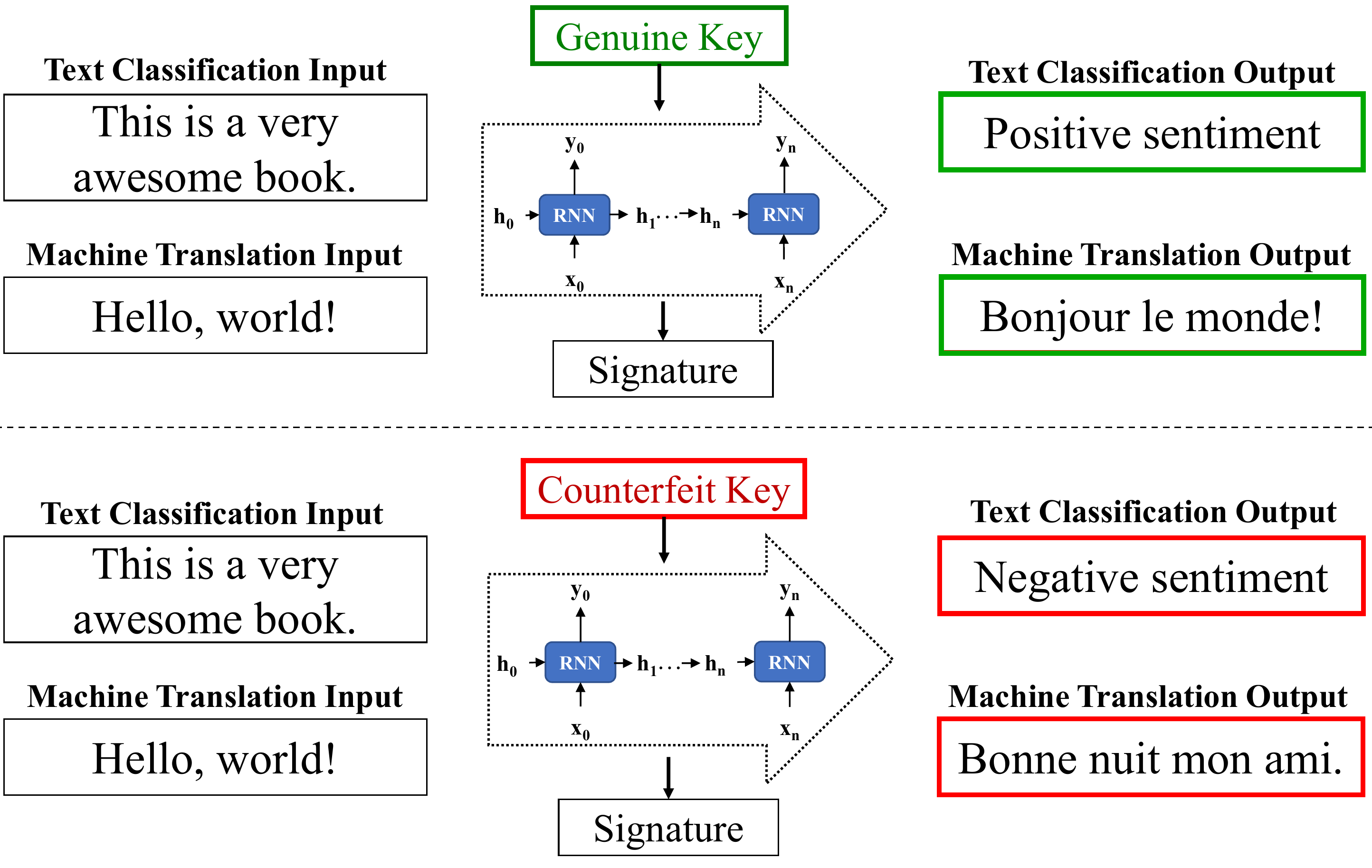}
\caption{Overview of our proposed IPR protection scheme in white/black box settings. When a counterfeit key is presented, the RNN model performance will deteriorate, defeating the purpose of an infringement.
}
\label{fig:overallprocess}
\vspace{-10pt}
\end{figure}

At the time of writing, there are already various DNN models protection schemes \cite{UchidaEmbeddingWatermark2017, ChenDeepsigns2018, ChenDeepmark2019, AdiBackdooring2018, ZhangProtecting2018, ErwanAdversarial2017, GuoWatermarking2018, FanDeepPassport2019, GanIPR2021}. In general, efforts to enforce IP protection on DNN can be categorized into two groups: i) \emph{white-box} (feature based) protection which embeds a watermark into the internal parameters of a DNN model (i.e. model weights) \cite{UchidaEmbeddingWatermark2017, ChenDeepmark2019, ChenDeepsigns2018}; and ii)  \emph{black-box} (trigger set based) protection which relies on specific input-output behaviour of the model through trigger sets (adversarial sample with specific labels) \cite{AdiBackdooring2018, ZhangProtecting2018, ErwanAdversarial2017, GuoWatermarking2018}. There are also protection schemes that utilize both white-box and black-box methods \cite{FanDeepPassport2019, GanIPR2021}. 

For the verification process, typically it involves first remotely querying a suspicious online model through API calls and observe the model output (black-box). If the model output exhibits a similar behaviour as to the owner embedded settings, it will be used as early evidence to identify a suspect. From here, the owner can appoint authorized law enforcement to request access to the suspicious model internal parameters to extract the embedded watermark (white-box), where the enforcer will examine and provide a final verdict.

\subsection{Problem Statement}
Recurrent Neural Network (RNN) has been widely used in various Natural Language Processing (NLP) applications such as text classification, machine translation, question answering etc. Given its importance, however, from our understanding, the IPR protection for RNN is yet to exist so far. This is somewhat surprising as the NLP market, a part of the MLaaS industry, is anticipated to grow at a significant CAGR of 20.2\% during the forecast period from 2021-2030. That is to say, the market is expected to reach USD 63 billion by 2030 \cite{mlmarket2022}. 

\subsection{Contributions}
The contributions of our work are twofold: 
\begin{enumerate}
\item We put forth a simple and generalized RNN ownership protection technique, namely the \emph{Gatekeeper} concept (Eqn. \ref{eqn:keygate}), that utilizes the endowment of RNN variant's cell gate to control the flow of hidden states, depending on the presented key (see Fig. \ref{fig:activation_main}); 
\item Extensive experimental results show that our proposed ownership verification (both in white-box and black-box settings) is \emph{effective} and \emph{robust} against removal and ambiguity attacks (see Table \ref{table:removalattack}) and at the same time, without affecting the model's overall performance on its original tasks (see Table \ref{table:performance}). 
\end{enumerate}

The proposed IPR protection framework is illustrated in Fig. \ref{fig:overallprocess}. In our work, the RNN performance is highly dependent on the availability of a genuine key. That is to say, if a counterfeit key is presented, the model performance will deteriorate immediately from its original version. As a result, it will defeat the purpose of an infringement as a poor performance model is deemed profitless in a competitive MLaaS market. 

\begin{figure*}[t]
\centering
\begin{subfigure}{0.49\linewidth}
  \centering
  \includegraphics[keepaspectratio=true, scale = 0.22]{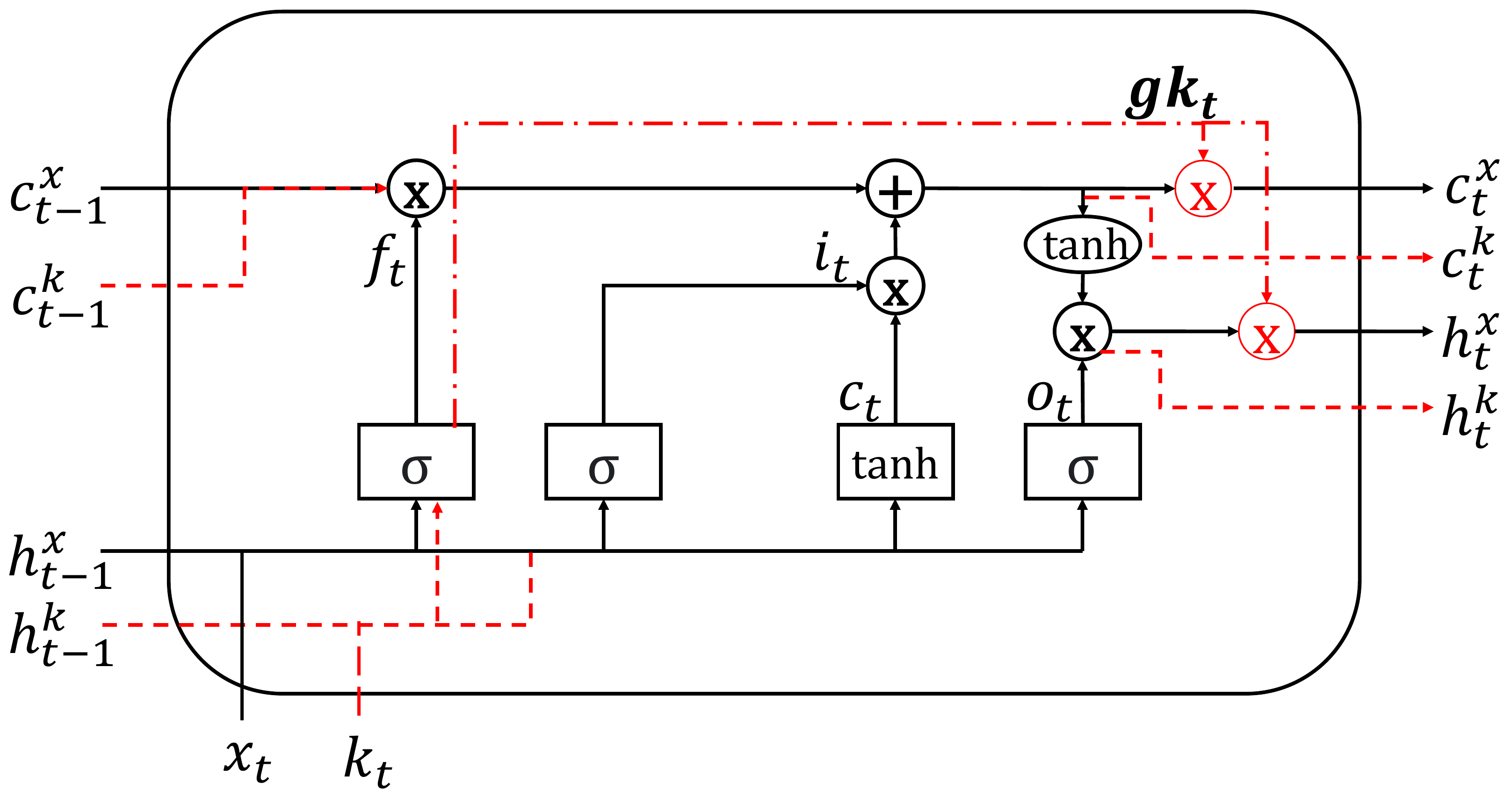}
  \caption{LSTM cell with Gatekeeper}
  \label{fig:lstm}
\end{subfigure}
 \hspace*{\fill}
\begin{subfigure}{0.49\linewidth}
  \centering
  \includegraphics[keepaspectratio=true, scale = 0.22]{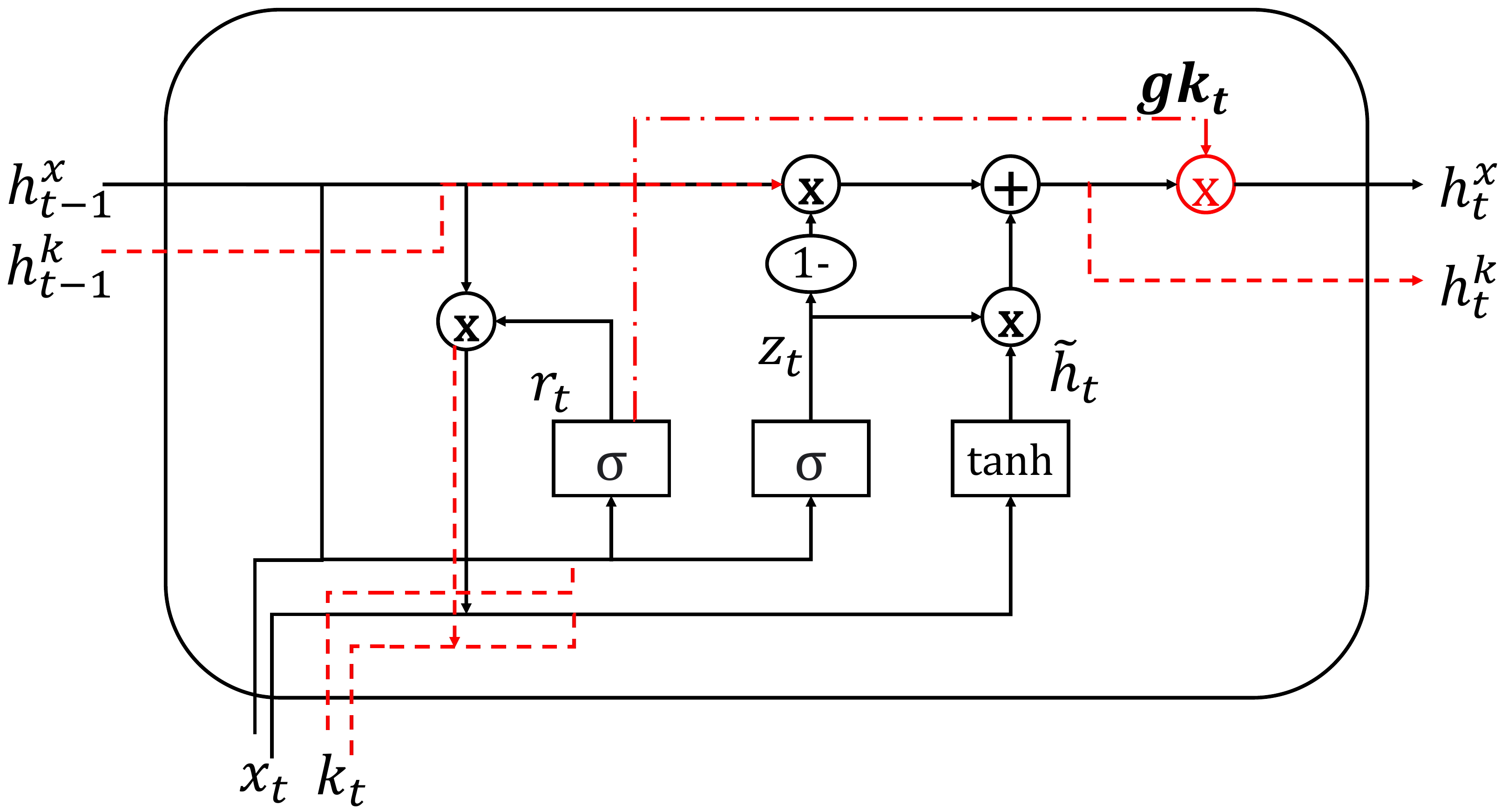}
  \caption{GRU cell with Gatekeeper}
  \label{fig:gru}
\end{subfigure}
\caption{Our proposed method in two major RNN variants: (a) LSTM; and (b) GRU. Solid lines denote the original RNN operation for each cell type. Dotted red lines delineate the proposed \emph{Gatekeeper}, which embeds a key recurrently with a new gate control manner, but without introducing extra weight parameters. Best viewed in colour.}
\label{fig:keygate}
\end{figure*}

\section{Related Work}
 \citet{UchidaEmbeddingWatermark2017} were the first to propose white-box protection to embed watermarks into CNN by imposing a regularization term on the weights parameters. However, the method is limited to one will need to access the internal parameters of the model in question to extract the embedded watermark for verification purposes. Therefore, \citet{QuanWatermark2021}, \citet{AdiBackdooring2018} and \citet{ErwanAdversarial2017} proposed to protect DNN models by training with  classification labels of adversarial examples in a trigger set so that ownership can be verified remotely through API calls without the need to access the model weights (black-box). In both black-box and white-box settings, \citet{GuoWatermarking2018, ChenDeepmark2019} and \citet{ChenDeepsigns2018} demonstrated how to embed watermarks (or fingerprints) that are robust to various types of attacks such as model fine-tuning, model pruning and watermark overwriting. Recently, \citet{FanDeepPassport2019} and \citet{ZhangPassport2020} proposed passport-based verification schemes to improve the robustness against ambiguity attacks. \citet{GanIPR2021} also proposed a complete IP protection framework for Generative Adversarial Network (GAN) by imposing an additional regularization term on all GAN variants. Other than that, \citet{PulkitRNNAdv2022} demonstrated how to generate adversarial examples by adding noise to the input of a speech-to-text RNN model in  black-box setting. Finally, \citet{NLGAPIHe2021} also proposed a protection method designed for language generation API by performing lexical modification to the original inputs in the black-box setting.

To the best of our knowledge, the closest work to ours is  \citet{lim2022protect}, applied on image captioning domain where a secret key is embedded into the RNN decoder for functionality-preserving. Although it looks similar to our idea, our proposed \emph{Gatekeeper} concept is a gate control approach rather than element-wise operation on the hidden states. That is to say, the embedded key in \citet{lim2022protect} is generated by converting a string into a vector; while in our work, the embedded key is a sequence of data similar to the input data. Furthermore, the key embedding operation in \citet{lim2022protect} method is a simple element-wise addition or multiplication between the fixed aforementioned vector and the RNN's hidden state. Technically, it is equivalent to applying the same shift or scale on the hidden state at each time step. In contrast, our proposed method adopts both the RNN weights and embedded key to calculate an activation \emph{recurrently} before performing the matrix multiplication on the hidden states at each time step (see Sec. \ref{sect:keyinfo}).

Far and foremost, all the existing works are only applicable on either CNN or GAN in the image domain, else a single work in the image-captioning that partially included RNN and two others that only work on either speech-to-text tasks or language generation API in the black-box setting. 
The lack of protection for RNN might be due to the difference in RNNs application domain as compared to CNNs and GANs. For example, \citet{UchidaEmbeddingWatermark2017} method could not be applied directly to RNNs due to the significant differences in both the input and output of RNNs as compared to CNNs. Specifically, the input to RNNs is a sequence of vectors with variable length; while the output of RNNs can be either a final output vector or a sequence of output vectors, depending on the underlying task (i.e. text classification or machine translation).
 
\section{RNN Ownership Protection}
 
Our idea for RNN models ownership protection is to take advantage of its existing recurrent property (sequence based), so that the information (hidden states) passed between timesteps will be affected when a counterfeit key is presented. Next, we will illustrate how to implement the \emph{Gatekeeper} concept to RNN cells, and then followed by how to verify the ownership via a new and complete ownership verification scheme. Note that, the \emph{Gatekeeper} concept uses a key $k$ which is a sequence of vectors similar to the input data $x$ (herein, the key will be a sequence of word embeddings. Please refer to Appx. \ref{appx:keymethod} for more details). Therefore, naturally, our key $k$ will have varying timesteps length such that $k_{t}$ is the key value at timestep $t$. 

We will demonstrate the proposed framework on two main RNN variants, namely LSTM \cite{HochLSTM1997} and GRU \cite{ChoGRU2014} and their respective bidirectional variants. However, one can easily apply it to other RNN variants such as Multiplicative LSTM \cite{BenMLSTM2016} and Peephole LSTM \cite{PeepLSTM2002}, etc. since the implementation is generic.

\subsection{Gatekeeper}
\label{sect:keyinfo}
As to the original design of RNN model, the choices and amount of information to be carried forward to the subsequent cells is decided by different combination of gates, depending on the RNN types. Inspired by this, we proposed the \emph{Gatekeeper} - a concept which learns to control the flow of hidden states, depending on the provided key (e.g. genuine key or counterfeit key). Technically, our \emph{Gatekeeper}, $gk_{t}$ is formulated as follows:
\begin{equation}
\label{eqn:keygate}
gk_{t} = \sigma (W_{ik}k_{t} + b_{ik} + W_{hk}h^{k}_{t-1} + b_{hk})
\vspace{-10pt}
\end{equation}
\begin{equation}
\label{eqn:keygate2}
h^{x}_{t} = gk_{t} \odot h^{x}_{t},\ \ \ c^{x}_{t} = gk_{t} \odot c^{x}_{t} \ \ \mbox{(for LSTM)}
\end{equation}
where $\sigma$ denotes sigmoid operation, $\odot$ is matrix multiplication, $k_{t}$ is the key value at timestep $t$, $h^{k}_{t-1}$ is the previous hidden state of the key, $h^{x}_{t}$ and $c^{x}_{t}$ (for LSTM) are the hidden state of the input, $x$. 

One of the key points of our Gatekeeper is it \emph{does not add weight parameters} to the protected RNN models as we chose to employ the original weights of a RNN to calculate the value of $gk_{t}$. That is, for LSTM cell, we use $W_{f}$ and $b_{f}$ \cite{HochLSTM1997} while for GRU cell, we use $W_{r}$ and $b_{r}$ \cite{ChoGRU2014} as $W_{k}$ and $b_{k}$, respectively. Note that the hidden state of a key at the next time step is calculated using the original RNN operation such that $h^{k}_{t} = R(k_{t}, h^{k}_{t-1})$ where $R$ represents the operation of a RNN cell. Fig. \ref{fig:keygate} outlines the architecture of RNN cell with our \emph{Gatekeeper} concept where Eqn. \ref{eqn:keygate} and Eqn. \ref{eqn:keygate2} are represented using the red dotted line, respectively. For a RNN model trained with key $k_{e}$, $\mathbb{N}[W, k_{e}]$, their inference performance $P$ of input, $x_{r}$ will depend on the running time key, $k_{r}$, such that
\begin{equation}
P(\mathbb{N}[W, k_{e}], x_{r}, k_{r}) = 
\begin{cases}
P_{k_{e}} & \mbox{if $k_{r} = k_{e}$} \\
\overline{P_{k_{e}}} & \mbox{otherwise}
\end{cases}
\label{eqn:inference}
\end{equation}

\begin{figure}[t]
\centering
\includegraphics[keepaspectratio=true, scale = 0.4]{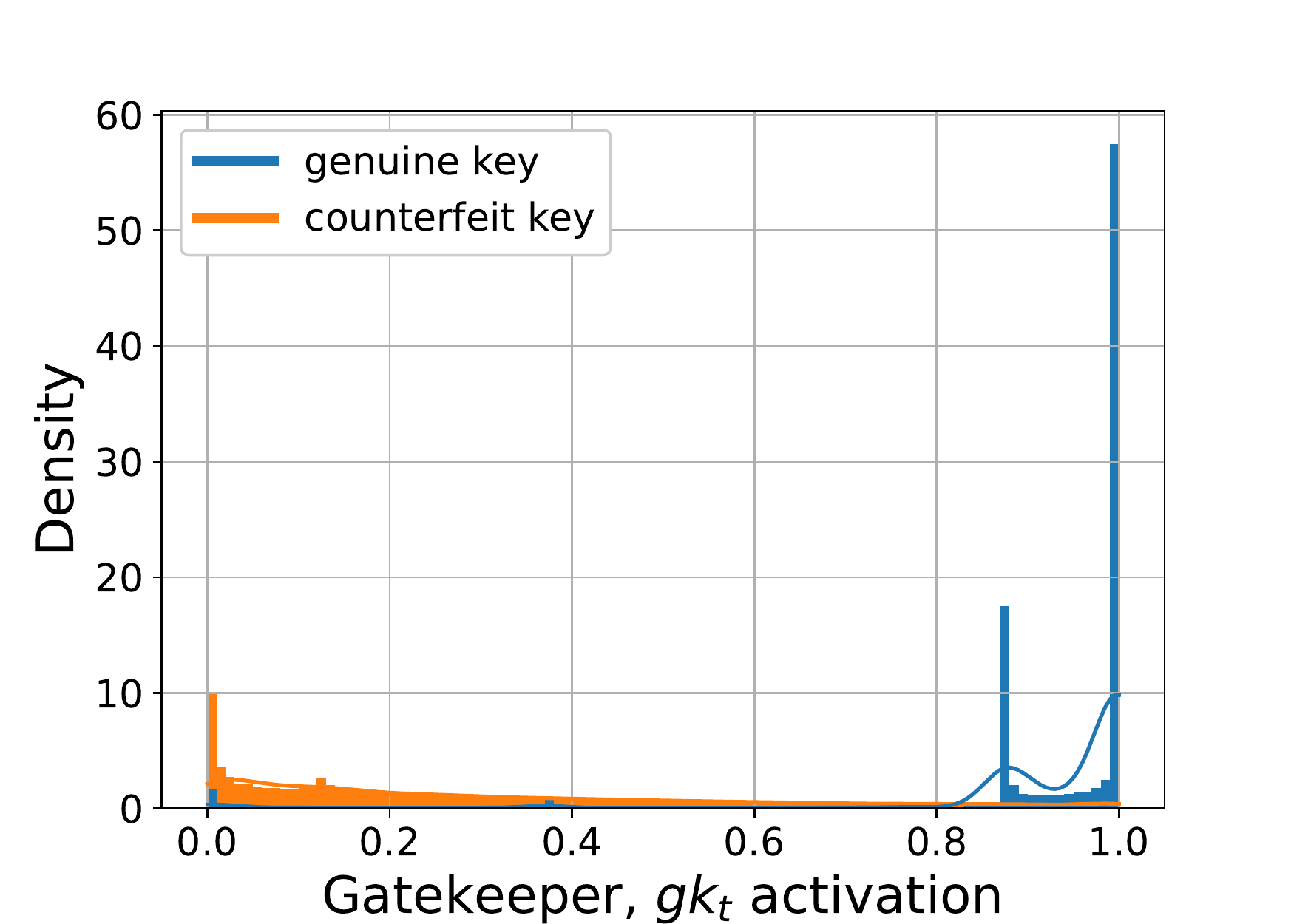}
\caption{Comparison of the Gatekeeper, $gk_{t}$ activation distribution when genuine/counterfeit key is employed.}
\label{fig:activation_main}
\end{figure}

That is to say if a genuine key is not presented $k_{r} \neq k_{e}$, the running time performance $\overline{P_{k_{e}}}$ will \emph{significantly deteriorate} because $gk_{t}$ is calculated based on an incorrect key. As an example, Fig. \ref{fig:activation_main} illustrates the distribution of $gk_{t}$ when the genuine and counterfeit keys are presented. It can be noticed that when the genuine key is presented, the $gk_{t}$ is mostly close to 1.0, thus allowing a proper flow of hidden states between time steps.  In contrast, when the counterfeit key is presented, $gk_{t}$ is miscalculated (most of the time is <1.0), thus \emph{disrupting} the flow of hidden states of input between time steps and causing the model to perform poorly from its original version. 

\subsubsection{Gatekeeper Sign as Digital Signature}
\label{sect:signinfo}
In order to further protect RNN models ownership, in particular from an insider threat (e.g. a former employee who establish a new business with all resources stolen from the original company), we can enforce the sign of the first hidden state of key $h^{k}_{0}$ to be either positive (+) or negative (-) signs as designated. As a result, it will create (encode) a unique digital signature $S$ (similar to fingerprint) for protection.  As an example, we can design $S$ to form a string - {\it ``This is the property of UniMalaya"} by encoding each ASCII character into its respective 8 bit code (See Appx. \ref{appx:signmethod} for more details). For this purpose, we adopted and modified the \emph{sign loss} regularization term proposed by \citet{FanDeepPassport2019} and add it to the combined loss such that:
\begin{equation}
L_{R}(h^{k}_{0}, S) = \sum_{i=1}^{N} max(\gamma - h^{k}_{0, i}s_{i}, 0) 
\label{eqn:signloss}
\end{equation}
\noindent where $S = {s_{1},\cdots, s_{N}} \epsilon \{-1, 1\}$ consists of the designated binary bits for $N$ hidden cell units in RNN and $\gamma$ is a positive control parameter (0.1 by default unless stated otherwise) to encourage the hidden state to have magnitudes greater than $\gamma$. 
Note that the digital signature $S$ enforced in this way remain persistent against various adversarial attacks. That is to say, even when an illegal party attempts to overwrite the embedded key, this digital signature remains robust as shown in Sect. \ref{sect:ambiguity}. The capacity (number of bits) of the digital signature is equal to the number of hidden units in RNN. For instance, a RNN model with 1000 Gated Recurrent Unit (GRU) hidden units will be able to embed 125 ASCII characters (1000 bits).

\subsection{Ownership Verification}
\label{verification}
In this section, we will discuss how to perform the ownership verification. With the introduction of \emph{Gatekeeper}, we have designed two new ownership verification schemes as follow. 
\begin{enumerate}
    \item \textbf{Private Ownership Scheme:} In this scheme, both the key and trigger set are embedded in the RNN model during the training phase. Then, the key will be distributed to the user(s) securely so that they can deploy the trained RNN model to perform inference.
    \label{item:privatescheme}
    
    \item \textbf{Public Ownership Scheme:} In this scheme, both the key and trigger set are embedded in the RNN model during the training phase as well, but the key will not be distributed to the user(s). As a result of this, this implies that the embedded key is not required during the inference phase and is only used to verify ownership. This is made possible with \emph{multi-task learning}. That is to say, technically, given a model $M$ protected with Gatekeeper $gk_{t}$, input data $X$, target $Y$ and a loss function $L$, first, we will calculate loss $L_{k}$ using $Y$ and output of model $M$ with $gk_{t}$ on $X$. Next, we will \emph{disable Gatekeeper temporarily} and calculate loss $L_{x}$ using $Y$ and output of model $M$ \emph{without $gk_{t}$} on $X$. The final loss is the summation of $L_{k}$ and $L_{x}$, which is then used to update the model's parameter at each training iteration. In a nutshell, the model learns to \emph{embed the key} and \emph{generate correct prediction without Gatekeeper} simultaneously.
    
    Algorithm~\ref{alg:process} shows the pseudo-code of Public Ownership Scheme via multi-task learning training, combined with the trigger sets protection. 
    \label{item:publicscheme}
    
\end{enumerate}

\paragraph{Trigger sets:} In this paper, we set the trigger sets, $\mathbf{T} \ni X_{t}, Y_{t}$ (see Table \ref{table:triggerexample}) for sequential tasks: (a) text classification and (b) machine translation as follows, but not limited to. For the text classification task, we randomly selected $t$ samples as the trigger set from the training dataset and shuffled their labels. Meanwhile for machine translation task, we investigated two different settings to create the trigger set: (i) randomly selected $t$ samples as the trigger set from the training dataset and shuffled their target translation; and (ii) create random sentences from the vocabulary $V$ of both source and target language as the trigger set. Empirically, both settings give similar performance. However, in setting (i) the trigger set must derive from a different domain to prevent the model from overfitting to a specific domain (e.g. training set = parliament speech, while trigger set = news commentary).  

\begin{table*}
  \caption{Examples of trigger set, $\mathbf{T}$ in text classification (TREC-6) and machine translation (WMT14 EN-FR) used in this paper. For text classification, the original labels are denoted in brackets. While for machine translation, the trigger output, $Y_t$ is constructed from the set of words from the target language vocabulary. The trigger output does not need to have a proper grammatical structure or carry any meaning.}
  \label{table:triggerexample}
  \centering
  \resizebox{0.8\textwidth}{!}{ 
  \begin{tabular}{lll}
    \toprule
    Tasks    & Trigger input, $X_t$     & Trigger output, $Y_t$ \\
    \midrule
    \multirow{3}{*}{Text classification} & When was Ozzy Osbourne born? & DESC (NUM) \\
                                         & What is ethology? & NUM (DESC) \\
                                         & Who produces Spumante? & LOC (HUM) \\
    \midrule
    \multirow{3}{*}{Machine translation} & Who are our builders? & Nous avons une gr{\^a}ce du Pape. \\
                                         & But I don't get worked up. & Je suis pour cette culture.  \\
                                         & Basket, popularity epidemics to & Desquels le constatons habillement \\
    \bottomrule
  \end{tabular}}
\end{table*}

\section{Experiment Results}
This section presents the empirical results of the proposed IPR protection framework for RNN models. Particularly, we will report results from the aspect of \emph{fidelity},  \emph{robustness}, \emph{secrecy} and \emph{time complexity} on two different tasks: i) text classification (TREC-6 \cite{LiTREC2002}); and ii) machine translation (WMT14 EN-FR \cite{WMT14}). Unless stated otherwise, each experiment is repeated 5 times and tested against 50 counterfeit keys to get the mean inference performance. Note that all the protected models presented in this section are protected with {\bf Public Ownership Scheme} and represented as follows: RNN$_{k}$ represents the protected model in the white-box settings, whereas RNN$_{kt}$ represents the protected model in both the white-box and black-box settings. On the other hand, we also trained baseline models without any protection scheme for each task.

\begin{algorithm}
\begin{algorithmic}[1]
\Function{Train}{$M$ w/ $gk_{t}$, $k$, $S$, $X$, $Y$, $X_{t}$, $Y_{t}$, $L$, $L_{R}$}
\ForAll{number of training iterations}
  \LineCommentAfter{sample $m$ batch of data from $X$, $Y$}
  \State $x_m$, $y_m$ = \Call{sample}{$m$, $X$, $Y$};
  \State $x_{nt}$, $y_{nt}$ = \Call{sample}{$n$, $X_{t}$, $Y_{t}$};
  \LineComment{concatenate $x_m$, $x_{nt}$ along first axis}
  \State $x$ = \Call{concat}{$x_m$, $x_{nt}$};
  \State $y$ = \Call{concat}{$y_m$, $y_{nt}$};
  \State Enable $gk_{t}$ in $M$;
  \State $L_{k}$ = $L(y, M(x, k))$;
  \State Disable $gk_{t}$ in $M$;
  \State $L_{x}$ = $L(y, M(x))$;
  \State $L_{r}$ = $L_{R}(S)$;
  \State $L_{total}$ = $L_{k} + L_{x} + L_{r}$;
  \LineComment{update parameters of $M$ using $L_{total}$ with backpropagation}
  \State \Call{UpdateParams}{$M$, $L_{total}$};
\EndFor
\EndFunction

\end{algorithmic}

\caption{Training step for Public Ownership Scheme}
\label{alg:process}
\end{algorithm}

\subsection{Experiment settings}
We chose the work by \citet{ChoGRU2014} and \citet{ZhouTRECbaseline2016} as the baseline models and followed the hyperparameters defined in their works for each task, i.e. machine translation on WMT14 EN-FR \cite{WMT14}, and text classification on TREC-6 \cite{LiTREC2002}. For machine translation task, we adopted a Seq2Seq model that comprises of an encoder and decoder with GRU layers similar to the baseline paper \cite{ChoGRU2014}. Please refer to Appx. \ref{appx:hyper} for complete information on the hyperparameters. In terms of metric evaluation, BLEU score \cite{PapineniBLEU2002} is used to evaluate the quality of the translation results.

\subsection{Fidelity}
The idea of fidelity refers to the degree to which a model reproduces the state and behaviour of a real world condition. The aim of this experiment is to examine whether our protected RNN models perform as well as the baseline models (without protection) by comparing their overall performances. As seen in both Table \ref{table:performance} and Table \ref{table:wmt14qualperfnew}, all the protected RNN models achieve an overall performance that is very similar to their respective baseline models. For instance, in TREC-6 dataset, the difference between BiGRU$_{k/kt}$ vs BiGRU is less than 2.5\% for all settings. A similar observation is also found on Seq2Seq$_{k/kt}$ for WMT14 EN-FR dataset. In summary, the introduction of our \emph{Gatekeeper} has \emph{minimal to no effect} on the original performance of the RNN model in their respective tasks. Please see Appx. \ref{appx:qualres} for more qualitative results.

\begin{table*}[t]
\caption{Comparison results for different protected RNN models where they are evaluated under 3 different scenarios: (i) w/o key = without key; (ii) w/ key = with genuine key; and (iii) $c$ key = with counterfeit key, in 2 different settings: (iv) Model$_k$ = white box; and (v) Model$_{kt}$ = white and black box. Original RNN models are in bold.} \label{table:performance}
\begin{subtable}{1\linewidth}
\caption{Performance on TREC-6} \label{table:trecperf}
\centering
\resizebox{0.9\linewidth}{!}{ 
   \begin{tabular}{l|c|ccc|ccc|c}
      \toprule
      & Train time & \multicolumn{3}{c|}{Test set}  & \multicolumn{4}{c}{Trigger set}  \\ 
      & (mins)    &   w/o key    & w/ key & $c$ key &      w/o key     & w/ key & $c$ key  & p-value \cite{NLGAPIHe2021} \\ 
      \midrule \midrule
      {\bf BiLSTM (baseline)} & {\bf 1.57} & {\bf 87.88} & - & - & - & -& - & {\bf > 10$^{\mathrm{-1}}$ } \\ 
      BiLSTM$_{k}$ (ours) & 6.53 & 86.71 & 86.92 & 76.03 $\downarrow$ & - & - & - & > 10$^{-1}$  \\ 
      BiLSTM$_{kt}$ (ours) & 6.61 & 86.16 & 86.21 & 75.78 $\downarrow$ & 100 & 99.81 & 44.79 $\downarrow$ & < 10$^{-10}$ \\ 
      \midrule
      {\bf BiGRU (baseline)} &  {\bf 1.60} & {\bf 88.48} & - & - & - & -& - & {\bf > 10$^{\mathrm{-1}}$ } \\ 
      BiGRU$_{k}$ (ours) & 6.34 & 87.46 & 87.64 & 84.11 $\downarrow$ & - & - & - & > 10$^{-1}$ \\ 
      BiGRU$_{kt}$ (ours) & 6.38 & 86.05 & 86.79 & 83.76 $\downarrow$ & 100 & 100 & 64.58 $\downarrow$ & < 10$^{-10}$ \\ 
      \bottomrule
   \end{tabular}}
\end{subtable}
\begin{subtable}{\linewidth}
\caption{Performance on WMT14 EN-FR} \label{table:wmt14perf}
\centering
\resizebox{0.9\linewidth}{!}{ 
   \begin{tabular}{l|c|ccc|ccc|c}
      \toprule
       & Train time & \multicolumn{3}{c|}{Test set}  & \multicolumn{4}{c}{Trigger set}  \\ 
       & (mins)    &  w/o key   & w/ key & $c$ key &     w/o key      & w/ key & $c$ key & p-value \cite{NLGAPIHe2021} \\ 
      \midrule \midrule
      {\bf Seq2Seq (baseline)} & {\bf 3062.83} & {\bf 29.33} & - & -  & - & - & -  &  {\bf > 10$^{\mathrm{-1}}$ } \\
      Seq2Seq$_{k}$ (ours) & 6090.78 & 29.60 & 29.74 & 14.92 $\downarrow$ & - & - & -  & > 10$^{-1}$ \\
      Seq2Seq$_{kt}$ (ours) & 6947.22 & 29.11 & 29.15 & 13.62 $\downarrow$ & 100 & 100 & 0.11 $\downarrow$ & < 10$^{-10}$ \\ 
      \bottomrule
   \end{tabular}}
\end{subtable}
\end{table*}

\begin{table*}
\caption{Qualitative results on WMT14 EN-FR. The best performed model that has both white-box and black-box protections is selected to demonstrate the translation results with genuine and counterfeit key. Best viewed in colour.} \label{table:wmt14qualperfnew}
\centering
\resizebox{\linewidth}{!}{
   \begin{tabular}{p{0.3\textwidth}|p{0.4\textwidth}|p{0.4\textwidth}|p{0.4\textwidth}}
      \toprule
      Input & Ground Truth & Translation with genuine key& Translation with counterfeit key $c$\\
      \midrule \midrule
      they were very ambitious . & ils étaient très ambitieux . & \textcolor{darkgreen}{ils ont très ambitieux .} & \textcolor{red}{elles ont $\langle$unk$\rangle$ $\langle$unk$\rangle$ en} \\
      \midrule
      the technology is there to do it . & la technologie est la pour le faire . & \textcolor{darkgreen}{la technologie est la pour le faire .} & \textcolor{red}{la technologie le la presente le $\langle$unk$\rangle$ .} \\
      \midrule
      to me , this is n't about winning or losing a fight . & pour moi, ceci n' est pas à propos de gagner ou de perdre une lutte . & \textcolor{darkgreen}{pour moi, ceci n' est pas à de gagner le perdre une lutte .} & \textcolor{red}{pour moi, n' est pas le à $\langle$unk$\rangle$ pour de de .}\\
      \midrule
      but that 's not all . & mais ce n' est pas tout . & \textcolor{darkgreen}{mais ce n' est pas tout .} & \textcolor{red}{mais cela n' est pas le à .} \\
      \bottomrule
   \end{tabular}}
\vspace{-5pt}
\end{table*}

\subsection{Verification}
\paragraph{Black-box:} In this setting, ownership can be verified by observing the model's output with our trigger set designed in Table \ref{table:triggerexample}, but not limited to. Table \ref{table:performance} shows that the accuracy/BLEU scores for all the protected models are high when the trigger input, $X_t$ with a genuine key is presented. Contrarily, the performance drops drastically; for instance, BiGRU$_{kt}$ drops from 100\% $\to$ 64.58\%. The owner can use this early evidence to identify a suspect quickly. Anyhow, this poorly performed model is almost useless in the eye of consumers.

Nonetheless, we also adopted another verification process as to \citet{NLGAPIHe2021}. For this, following the original work \cite{NLGAPIHe2021}, p-value \cite{rice2007mathematical} was chosen as the evaluation metric. Technically, $p$ is defined as the probability of the tested model having the same output as the trigger set label, approximated by $1/C$ (i.e. $C$ is the number of possible classes for the text classification task). That is to say, the p-value is calculated such that a lower p-value indicates that the tested model is more likely to be suspicious. Table \ref{table:performance} shows that BiLSTM$_{kt}$, BiGRU$_{kt}$ and Seq2Seq$_{kt}$ have a much smaller p-value when compared to their respective baseline models. Note that BiLSTM$_k$, BiGRU$_k$ and Seq2Seq$_k$ are protected in white-box settings only and therefore exhibit similar p-value as to their respective baseline models.

\paragraph{White-box:} In this setting, we can verify ownership by comparing the model performance, using the genuine key from the owner against the counterfeit key $c$ from the suspect. Table \ref{table:performance} shows that when a genuine key is used, the protected models always achieve similar performance to their respective baseline models. In contrast, when a counterfeit key $c$ is used, we can observe a drop in the performance across all the protected RNN models. For instance, the BLEU score of Seq2Seq$_{kt}$ drops from 29.15 $\to$ 13.62 (almost 50\% drops). Qualitatively, a similar observation is also noticed in Table \ref{table:wmt14qualperfnew} for the machine translation task. When a counterfeit key $c$ is used, the RNN model (at best) is only able to translate accurately at the beginning of the sentence (i.e. {\it la technologie}), but the translation quality quickly deteriorated towards the end of the sentence (i.e. {\it le la presente le <unk>}). 

\begin{table}[t]
\caption{Robustness of protected RNN model (in bold) against removal attacks (i.e. fine-tuning and overwriting). All metrics reported herein are the performance with genuine key.} \label{table:removalattack}
\begin{subtable}{1\columnwidth}
\caption{Robustness on TREC-6} \label{table:trecrem}
\centering
\resizebox{0.9\textwidth}{!}{
   \begin{tabular}{l|ccc}
      \toprule
      & Test set & Trigger set & Digital Sign. \\ 
      \midrule \midrule
      {\bf BiLSTM$_{kt}$} & {\bf 86.21} & {\bf 99.81} & {\bf 100}\\ 
      Fine-tuning & 86.56 & 98.77 & 100\\
      Overwriting & 85.91 & 98.08 & 100\\
      \midrule
      {\bf BiGRU$_{kt}$} & {\bf 86.79} & {\bf 100} & {\bf 100}\\
      Fine-tuning & 86.69 & 99.23 & 100\\
      Overwriting & 86.02 & 98.08 & 100\\
      \bottomrule
   \end{tabular}}
\end{subtable}

\begin{subtable}{1\columnwidth}
\caption{Robustness on WMT14 EN-FR} \label{table:wmt14rem}
\centering
\resizebox{0.9\textwidth}{!}{
  \begin{tabular}{l|ccc}
      \toprule
      & Test set & Trigger set & Digital Sign. \\ 
      \midrule \midrule
      {\bf Seq2Seq$_{kt}$} & {\bf 29.15} & {\bf 100} & {\bf 100}\\
      Fine-tuning & 29.51 & 100 & 100\\
      Overwriting & 29.04 & 100 & 100 \\
      \bottomrule
  \end{tabular}}
\end{subtable}
\vspace{-1pt}
\end{table}

\subsection{Robustness against removal attacks}
In this section, we examine the robustness of our proposed Gatekeeper when an illegal party attempts to remove the embedded key through common model modification methods such as model pruning and fine-tuning.

\begin{figure}[t]
\centering
\begin{subfigure}{0.49\linewidth}
  \centering
  \includegraphics[keepaspectratio=true, width=\linewidth]{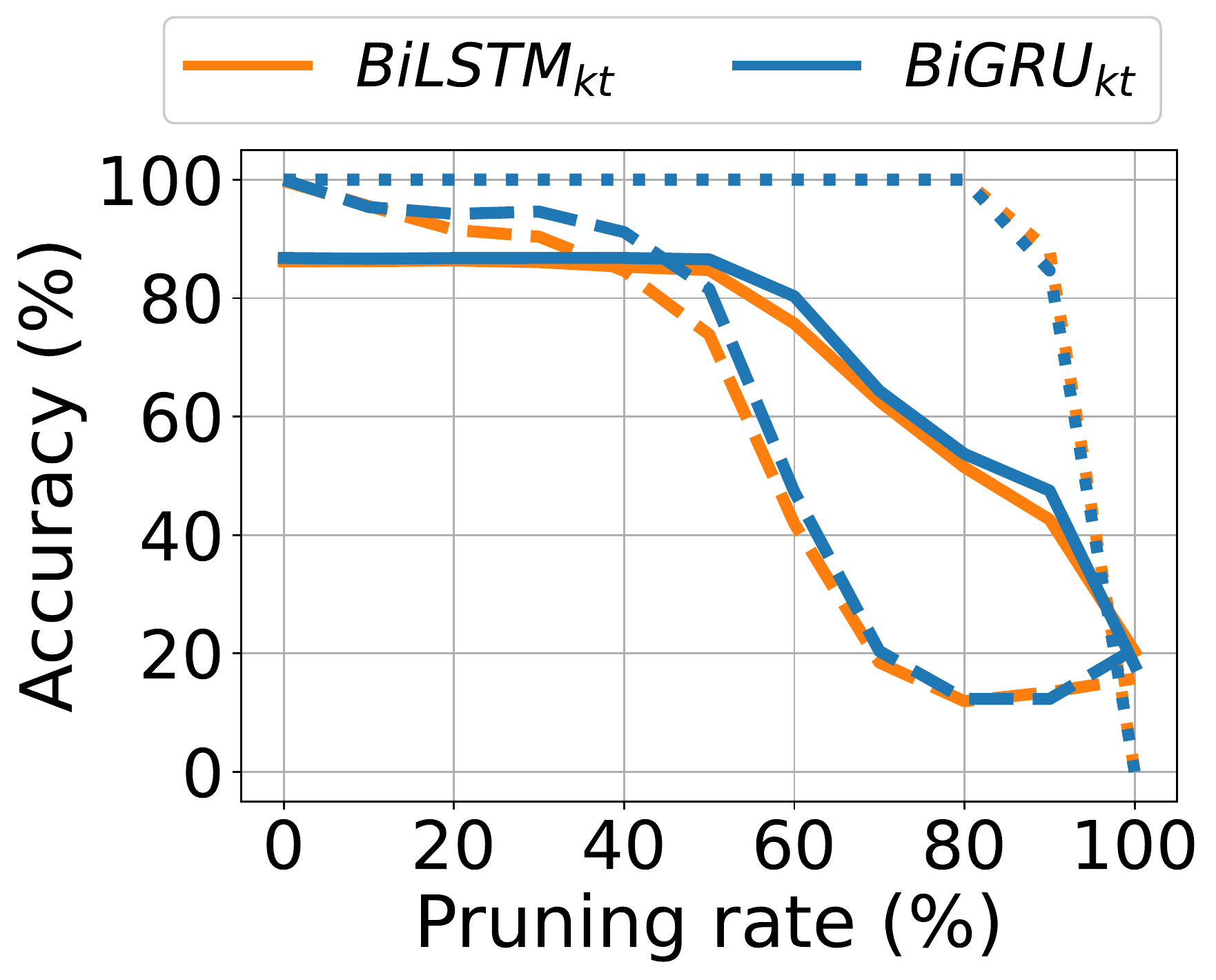}
  \caption{TREC-6}
  \label{fig:trecpruning}
\end{subfigure}
\hfill
\begin{subfigure}{0.49\linewidth}
  \centering
  \includegraphics[keepaspectratio=true, width=\linewidth]{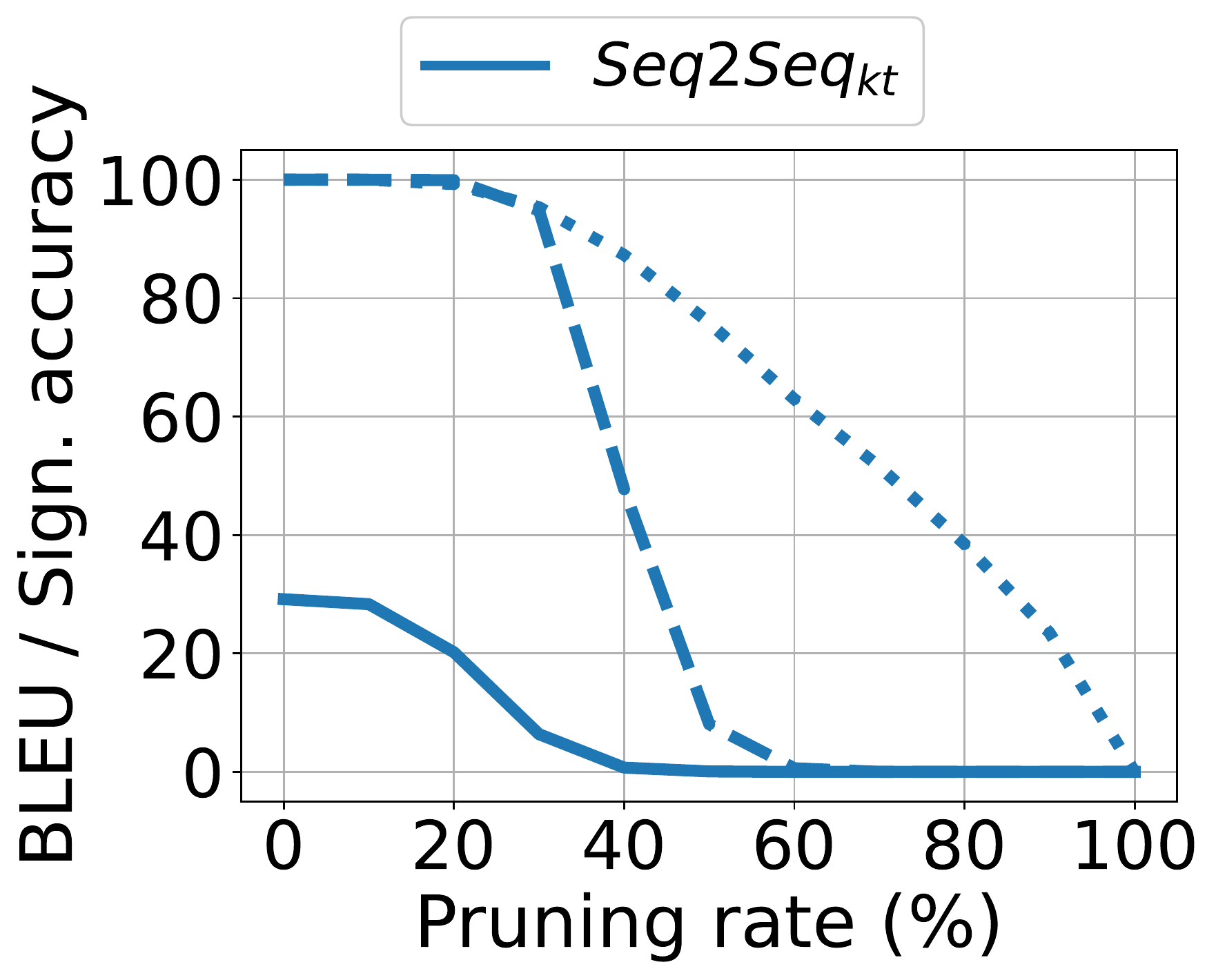}
  \caption{WMT14 EN-FR}
  \label{fig:wmtpruning}
\end{subfigure}
\caption{Robustness of the protected RNN models on test set (solid line), trigger set (dashed line) and digital signature (dotted line) against different pruning rates. Best viewed in colour.}
\label{fig:pruning}
\vspace{-10pt}
\end{figure}

\paragraph{Model Pruning}
This is a common model modification technique to remove redundant parameters in the deep learning model \cite{Pruning2016}. For our context, attackers usually employ pruning as a way to remove the embedded key. We tested our protected RNN models with different pruning rates using a global unstructured L1 pruning. In Figure \ref{fig:pruning}, we can observe that for both BiLSTM$_{kt}$ and BiGRU$_{kt}$ (see Fig. \ref{fig:trecpruning}) even at the point where 60\% of the parameters were pruned (in both test set and trigger set), the digital signature accuracy is still intact near to 100\% for ownership protection. However, one can also observe that both the protected RNN models' accuracy have dropped around 10\% - 20\% at this stage. As for the translation task (Fig. \ref{fig:wmtpruning}), at only 20\% of the parameters are pruned, BLEU score of Seq2Seq$_{kt}$ has already dropped by almost 30\%, yet the digital signature accuracy is still maintained at 100\%. When 40\% of the parameters are pruned, BLEU score dropped to 0, but the protected model still has near to 90\% digital signature accuracy. Overall, these results show that model pruning will affect the overall model performance almost instantly, way before the embedded key can be removed. As a summary, our proposed work is robust against model pruning.

\paragraph{Fine-tuning}
Here, we simulate an attacker that attempts to remove the embedded key by fine-tuning a stolen model with a new dataset. In short, the host model is initialized using the trained weights with the embedded key, then it is fine-tuned without the presence of the key, trigger set and regularization terms, i.e. $L_{R}$. In Table \ref{table:removalattack}, we can observe 100\% digital signature accuracy is detected for the ownership protection when the model is fine-tuned. Then, when the genuine key is presented to the fine-tuned model, all models have comparable performance on both test and trigger sets compared to the stolen model. Therefore, the proposed Gatekeeper and digital signature work together have provided a robust protection against fine-tuning.

\paragraph{Overwriting}
Here, we simulate an attacker who knows how the RNN model is protected, he/she can attempts to embed a new key, $\overline{k}$ into the trained model using the same method as detailed in Sect. \ref{sect:keyinfo}. In Table \ref{table:removalattack}, we can observe digital signature accuracy = 100\%, even when the protected model is overwritten with a new key. Then when inferencing using the original genuine key, most of the protected models' performance dropped slightly (less than 1\%). This confirms that it is hard to remove the embedded key and digital signature by overwriting it with new keys. However, this indirectly introduces an \emph{ambiguous situation} where there will be multiple keys (e.g. the original genuine key and overwritten new key) that satisfy the key verification process as denoted in Sect. \ref{verification}. To resolve this, we will show next how to employ digital signature $S$ (Sec. \ref{sect:signinfo}) to verify ownership.

\subsection{Resilience against ambiguity attacks}
\label{sect:ambiguity}

\begin{figure}[t]
\centering
\begin{subfigure}{0.49\linewidth}
  \centering
  \includegraphics[keepaspectratio=true, width=\linewidth]{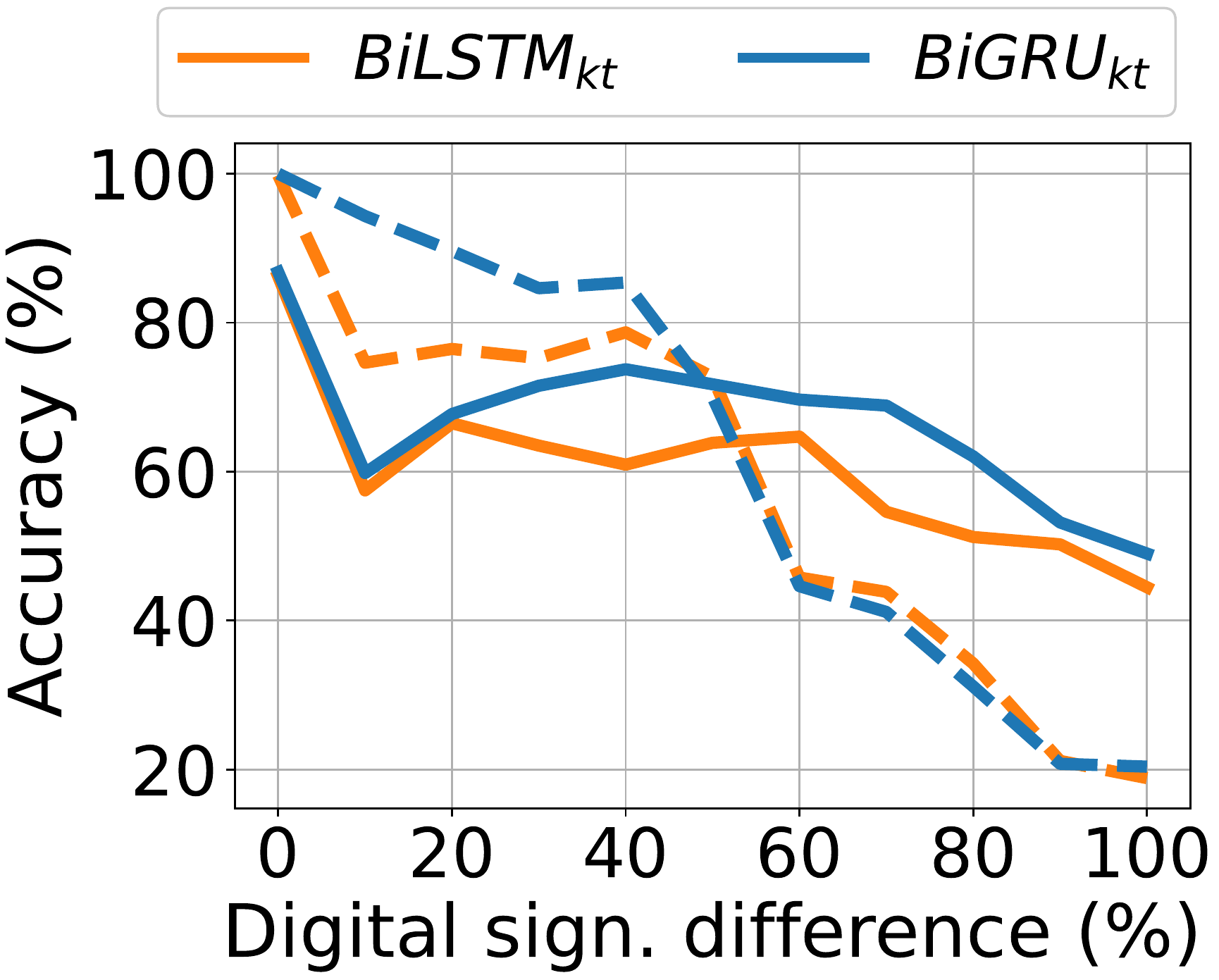}
  \caption{TREC-6}
  \label{fig:trecsignflip}
\end{subfigure}
\hfill
\begin{subfigure}{0.49\linewidth}
  \centering
  \includegraphics[keepaspectratio=true, width=\linewidth]{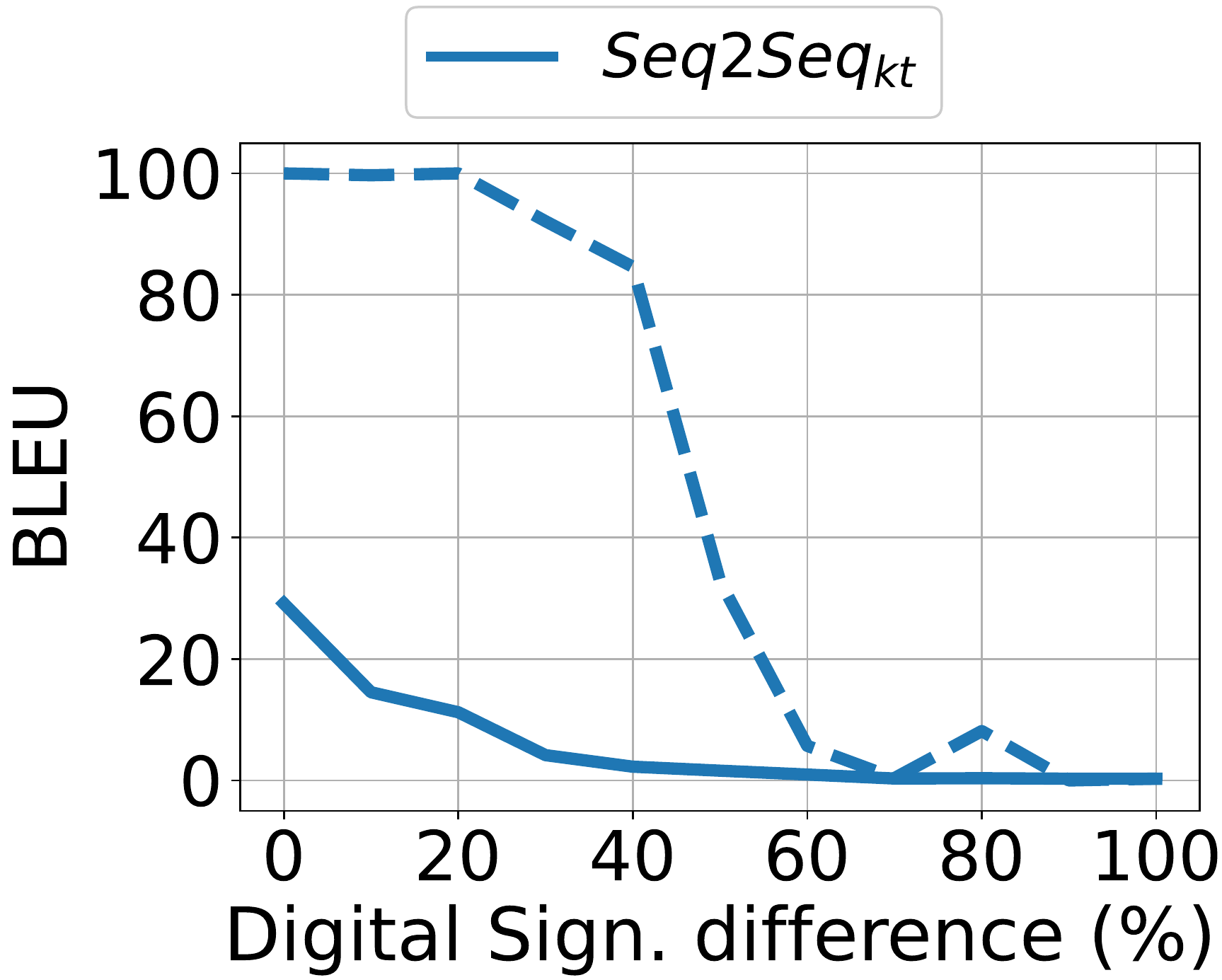}
  \caption{WMT14 EN-FR}
  \label{fig:wmtsignflip}
\end{subfigure}
\caption{Classification accuracy for classification tasks and BLEU score for translation task on test set (solid line) and trigger set (dashed line) when different percentage (\%) of the digital signature $S$ is being modified/compromised. Best viewed in colour.}
\label{fig:signflip}
\end{figure}

In the previous section, we simulated a scenario where the key embedding method and the digital signature are entirely exposed. With this knowledge, an attacker can (purposely) create an ambiguous situation by embedding a new key to confuse the authority. Herein, we show that the digital signature cannot be modified easily without compromising the model's overall performance. Figure \ref{fig:signflip} shows an example that when 40\% of the signs are being modified: for text classification task on TREC-6 (Fig. \ref{fig:trecsignflip}), the protected model's accuracy drops from 86.21\% $\to$ 60.93\% (for the test set in BiLSTM$_{kt}$); as for the translation task on WMT14 EN-FR, (Fig. \ref{fig:wmtsignflip}), the BLEU score drops from 29.15 $\to$ 2.27 (more than 90\% drop in the test set). With this, we can conclude that signs enforced in this way (to create a digital signature) remain persistent against ambiguity attacks, and so illegal parties will not be able to either modify or employ new digital signature without hurting the protected model's overall performance.

\subsection{Secrecy}
\begin{figure}[t]
\centering
\begin{subfigure}{0.49\linewidth}
  \centering
  \includegraphics[keepaspectratio=true, width=\linewidth]{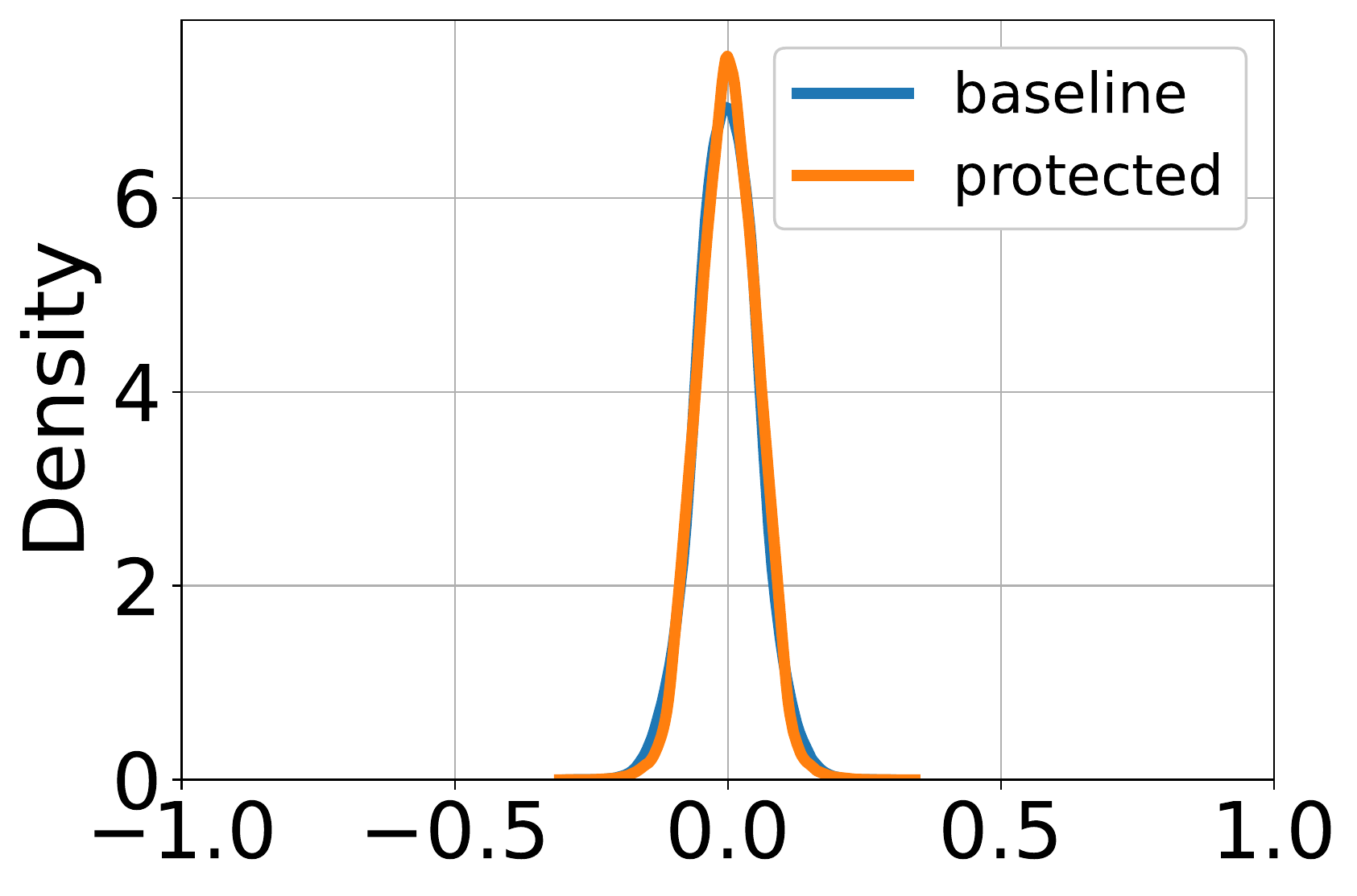}
  \caption{TREC-6}
  \label{fig:trecweight}
\end{subfigure}
\hfill
\begin{subfigure}{0.49\linewidth}
  \centering
  \includegraphics[keepaspectratio=true, width=\linewidth]{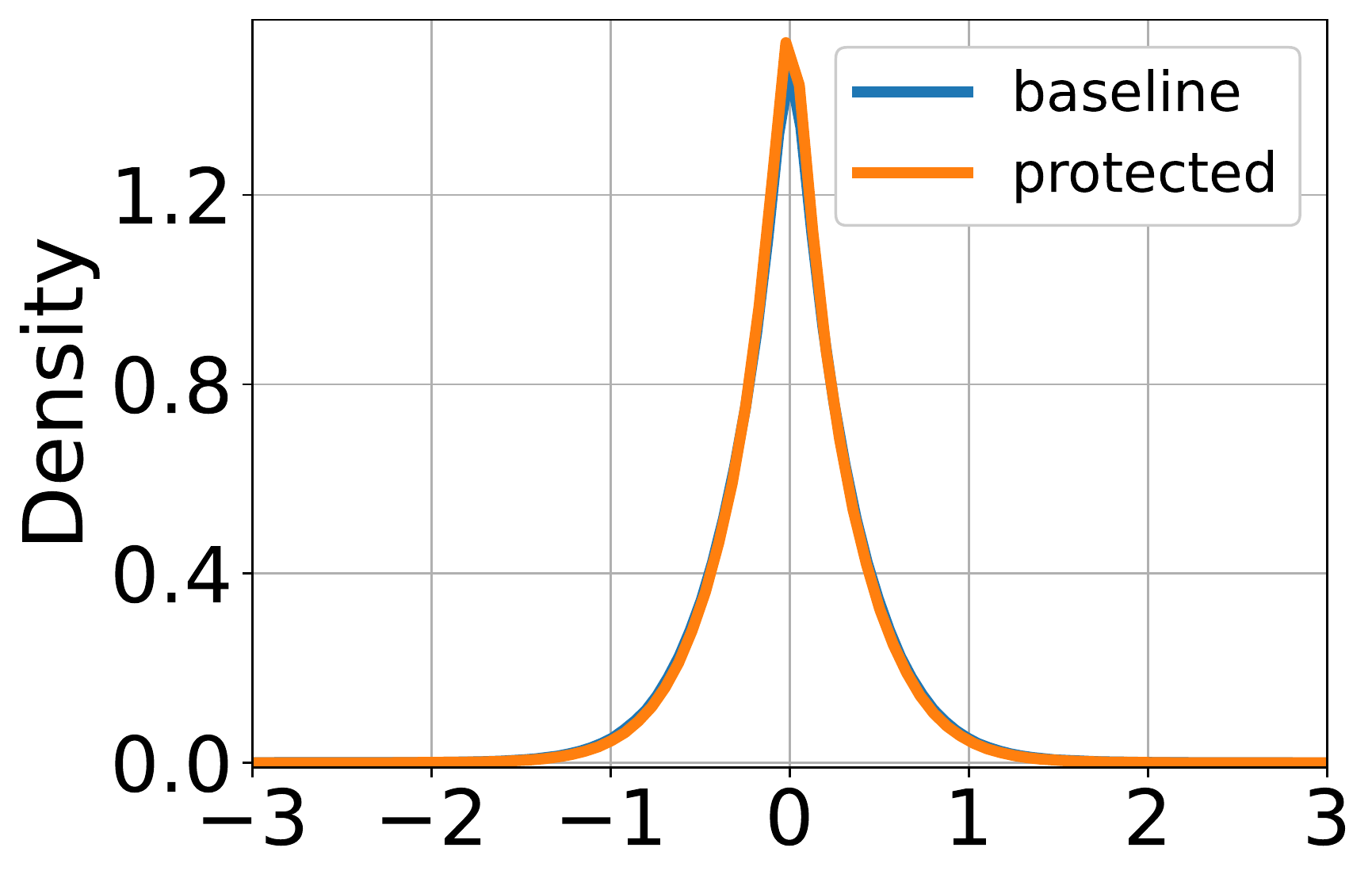}
  \caption{WMT14 EN-FR}
  \label{fig:wmtweight}
\end{subfigure}
\caption{Comparison of the weight distribution between baseline and protected RNN layer. Best viewed in colour.}
\label{fig:weight}
\end{figure}

Secrecy \cite{LiSurvey2021} is one of the requirements for watermarking techniques where the embedded watermark should be \emph{undetectable} and \emph{secret} to prevent unauthorized parties from being detecting it. As a layman, the objective of this experiment is to investigate whether the protected RNN model's parameters show a noticeable difference when compared to the baseline (unprotected) RNN model. Fig. \ref{fig:weight} shows the weight distribution of the protected RNN model against the baseline RNN model where it is observed that the weight distribution of the protected RNN layers (represented with orange colour) is identical to the baseline (represented in blue colour).

\begin{table*}
\caption{Results on SeqMNIST dataset for different protected RNN models evaluated under 3 different scenarios: (i) w/o key = without key; (ii) w/ key = with genuine key; and (iii) $c$ key = with counterfeit key, in 2 different settings: (iv) Model$_k$ = white box; and (v) Model$_{kt}$ = white and black box. Original RNN models are in bold.}
\label{table:mnistperfmain}
\centering
\resizebox{0.9\linewidth}{!}{
   \begin{tabular}{l|c|ccc|ccc|c}
      \toprule
      & Train time & \multicolumn{3}{c|}{Test set}  & \multicolumn{4}{c}{Trigger set} \\ 
       & (mins)    &  w/o key   & w/ key & $c$ key &     w/o key      & w/ key & $c$ key & p-value \cite{NLGAPIHe2021}  \\ 
      \midrule \midrule
      {\bf LSTM (baseline)} & {\bf 4.86} & {\bf 98.38} & - & - & - & -&- & {\bf > 10$^{\mathrm{-1}}$ } \\
      LSTM$_{k}$ (ours) & 18.85 & 98.36 & 98.37 &18.36 $\downarrow$ & - & - & - & > 10$^{-1}$\\
      LSTM$_{kt}$ (ours) & 19.53 & 98.17 & 98.18 &18.37 $\downarrow$ & 100 & 99.80 &6.51 $\downarrow$  & < 10$^{-10}$\\ 
      \midrule
      {\bf GRU (baseline)} & {\bf 4.74} & {\bf 98.36} & -& -  & - & - &- & {\bf > 10$^{\mathrm{-1}}$ }\\ 
      GRU$_{k}$ (ours) & 17.66 & 98.30 & 98.30 &22.68 $\downarrow$ & - & - &- & > 10$^{-1}$\\ 
      GRU$_{kt}$ (ours) & 18.69 & 97.97 & 97.95 &21.15 $\downarrow$ & 99.80 & 99.80 & 9.57 $\downarrow$ & < 10$^{-10}$\\ 
      \bottomrule
   \end{tabular}}
 \vspace{-1pt}
\end{table*}

\subsection{Time complexity}

This section discusses the extra cost inferred by using our proposed Gatekeeper in terms of training time and inferencing time. Table \ref{table:performance} shows the total training time (in minutes) of the protected RNN models, using Tesla P100 GPU. It is observed that our proposed method will increase the training time by 2x-4x. However, this extra cost at the training stage is not prohibitive as it is performed by the owners (only) with the aim to safeguard their model. Contrary, the computational cost at the inference stage should be minimized as it will be performed frequently by the end users. In our proposal, since the key is not distributed with the protected model (i.e Public Ownership Scheme),  there is no additional computational cost during the inference stage.

\section{Cross Domain Application}
In addition to the NLP domain, to show the generalizability of Gatekeeper, we also applied our proposed framework to the image domain, specifically in the task of sequential image classification. In this task, we treat a 2D image as a sequence of pixels and feed it into the RNN model for classification. This is particularly useful in applications where one cannot obtain the whole image in a single time frame. SeqMNIST \cite{QuocSeqMnist2015} is a variant of MNIST where the sequence of image pixels representing the handwritten digit images is classified into 10 digit classes. For the trigger sets, we follow the work by \citet{AdiBackdooring2018}, where we randomly select images from the training dataset and shuffle their labels. We chose \citet{QuocSeqMnist2015} as the baseline model and followed their hyperparameters exactly as a fair comparison. 

Quantitatively, as seen in Table \ref{table:mnistperfmain}, we achieve similar outcomes in the NLP domain. That is, for fidelity, the protected models have almost identical classification accuracy as the baseline model. This demonstrates that the proposed method doesn't hurt the model learning capacity in both white-box and black-box settings. Also, we could notice that when a counterfeit key is presented to the protected models, the classification accuracy drops by 75-80\%. As an example, for the white-box setting, the LSTM$_{kt}$ accuracy drops from 98.18\% $\to$ 18.37\%, while for the trigger set, its accuracy drops from 99.80\% $\to$ 6.51\% when a counterfeit key is presented. Please see Appx. \ref{appx:seqmnist} for more results.

\section{Conclusion and Future Works}
This paper demonstrates a simple but effective IPR protection method with complete and robust ownership verification scheme for RNNs in both white-box and black-box settings. The formulation of the \emph{Gatekeeper} is generic and can be applied to other variants of RNN directly. Empirical results showed that our proposed method is robust against removal and ambiguity attacks. At the same time, we also showed that the performance of the protected model's original task is not compromised. Future works are needed to ensure that the proposed \emph{Gatekeeper} is fully protected against overwriting attacks that introduce an ambiguous situation by embedding a new key simultaneously. 

\section*{Acknowledgement}
This research is supported by the 2021 SATU Joint Research Grant under grant no. ST041-2021 from Universiti Malaya; and the GPUs used for this research was donated by the NVIDIA Corporation. 

\section{Broader Impact}
Our proposed ownership protection framework aims to protect the IPR of RNN model. To compete with each other and gain business advantage, a large number of resources/budgets are continually being invested by giant and/or startup companies to develop new DNN models. Hence, we believe it is vital to protect these inventions from being abused, stolen or plagiarized. We believe that nobody with genuine intention will be harmed by this work. In the worst case scenario where if our proposed work fails to protect the RNN model; it just reflects the current status of RNN model as from our understanding, there is yet initiative of the IPR protection for RNN. In short, the ownership verification for RNNs will bring benefits to society by providing technical solutions to reduce plagiarism in deep learning and thus, lessen wasteful lawsuits and secure business advantages in an open market.

\small{
  \bibliographystyle{acl_natbib}
  \bibliography{anthology,custom}
}

\clearpage
\appendix
\section{Appendix}

\subsection{Hyperparameters}
\label{appx:hyper}
Table \ref{table:hyper} summarizes all the hyperparmeters used in the experiments.

\begin{table}[hp]
  \caption{Hyperparameters used in each tasks.}
  \label{table:hyper}
  \centering
  \resizebox{\linewidth}{!}{
  \begin{tabular}{lccc}
    \toprule
    Hyperparameter & TREC-6 & WMT14 EN-FR \\
    \midrule
    Vocabulary size & - & 15000 \\
    Max sentence length & 30 & 15 (EN) / 20 (FR) \\
    RNN hidden units & 300 & 1000\\
    Embedding dimension & 300 & 300 \\
    Batch size & 10 & 256 \\
    Bidirectional & Yes & No \\
    Optimizer & Adam\cite{Adam2015} & Adam \\
    \bottomrule
  \end{tabular}}
\end{table}

\begin{table}[t]
  \caption{Example of hidden state output $h^{k}_{0}$ and their respective sign (+/-) from LSTM$_{kt}$ when we embed digital signature $S$=\{\textit{private signature goes here}\}}
  \label{table:samplesign}
  \centering
  \resizebox{0.9\linewidth}{!}{
  \begin{tabular}{cc|cc}
    \toprule
    Hidden state $h^{k}_{0}$ & Sign (+/-) & ASCII code & Character\\
    \midrule \midrule
    -0.1939 & -1 & \multirow{8}{*}{112} & \multirow{8}{*}{p}\\
    0.1820 & 1 \\
    0.2064 & 1 \\
    0.1648 & 1\\
    -0.1795 & -1\\
    -0.1670 & -1\\
    -0.1778 & -1\\
    -0.1711 & -1\\
    \midrule \midrule
    -0.2059 & -1 & \multirow{8}{*}{114} & \multirow{8}{*}{r}\\
    0.1685 & 1 \\
    0.1767 & 1 \\
    0.1876 & 1\\
    -0.1996 & -1\\
    -0.1997 & -1\\
    0.1882 & 1\\
    -0.1655 & -1\\
    \midrule \midrule
    -0.1657 & -1 & \multirow{8}{*}{105} & \multirow{8}{*}{i}\\
    0.1838 & 1 \\
    0.2144 & 1 \\
    -0.1840 & -1\\
    0.1652 & 1\\
    -0.1818 & -1\\
    -0.2118 & -1\\
    0.1673 & 1\\
    \midrule \midrule
    -0.2330 & -1 & \multirow{8}{*}{118} & \multirow{8}{*}{v}\\
    0.1882 & 1 \\
    0.1740 & 1 \\
    0.1909 & 1\\
    -0.1963 & -1\\
    0.1868 & 1\\
    0.1882 & 1\\
    -0.1951 & -1\\
    \bottomrule
  \end{tabular}}
\end{table}

\subsection{Qualitative Results} 
\label{appx:qualres}
Table \ref{table:wmt14qualperfnew} and \ref{table:trecqualperf} show examples of incorrect predictions when a counterfeit key is embedded into the recurrent neural network (RNN) model during inference phase. For classification tasks (i.e. TREC-6  \cite{LiTREC2002}), Table \ref{table:trecqualperf} shows that when a counterfeit key is used, the RNN model gets confused between similar classes, i.e. DESC and ENTY for TREC-6. Meanwhile, for machine translation task (i.e. WMT14 EN-FR \cite{WMT14}), Table \ref{table:wmt14qualperfnew} demonstrates the translation results when a genuine key is used against a counterfeit key. It is observed that when a counterfeit key is used, the RNN model can still somehow translate accurately at the beginning of the sentence, but the translation quality quickly deteriorates toward the end of the sentence. This is in line with our idea and design of Gatekeeper where the information (hidden state) passed between timesteps would be disrupted with a counterfeit key and the output of RNN would deviate further from the ground truth the longer the timesteps are.

\begin{table*}[t]
\caption{Qualitative results on TREC-6. The best-performed model that has both white-box and black-box protections is selected to demonstrate the classification results with genuine and counterfeit keys.} \label{table:trecqualperf}
\centering
\resizebox{0.8\linewidth}{!}{
   \begin{tabular}{p{0.35\textwidth}|p{0.15\textwidth}|p{0.15\textwidth}|p{0.15\textwidth}}
      \toprule
      Input & Ground Truth & Prediction with & Prediction with \\
       &  & genuine key & counterfeit key \\
      \midrule \midrule
      What is Mardi Gras ? & DESC & \textcolor{darkgreen}{DESC} & \textcolor{red}{ENTY} \\
      \midrule
      What date did Neil Armstrong land on the moon ? & NUM &\textcolor{darkgreen}{ NUM} & \textcolor{red}{DESC} \\
      \midrule
      What is New York 's state bird ? & ENTY & \textcolor{darkgreen}{ENTY} & \textcolor{red}{DESC} \\
      \midrule
      How far away is the moon ? & NUM & \textcolor{darkgreen}{NUM} & \textcolor{red}{LOC} \\
      \midrule
      What strait separates North America from Asia ? & LOC & \textcolor{darkgreen}{LOC} & \textcolor{red}{ENTY }\\
      \bottomrule
   \end{tabular}}
\end{table*}

\subsection{Methods to generate key}
\label{appx:keymethod}
Three types of methods to generate key have been investigated in our work: 
\begin{itemize}
    \item \emph{random patterns}, elements of key are randomly generated from a uniform distribution between [-1, 1]. For natural language processing (NLP) task, a sequence of random word embedding can be used.
    \item \emph{fixed key}, one key is created from the input domain and fed through the trained RNN model with the same architecture to collect its corresponding features at each layer. The corresponding features are used in the Gatekeeper. For NLP task, a sentence from the input language domain is used as key.
    \item \emph{batch keys}, a batch of $K$ keys similar to above are fed through the trained RNN model with the same architecture. Each $K$ features is used in the Gatekeeper, and their mean value is used to generate the final Gatekeeper activation.
\end{itemize}
In the \emph{batch keys} method, the number of possible key combination is $(K \times l)^{V}$ where $K$ is the number of keys used, $l$ is the length/time step of key and $V$ is the vocabulary size. This make it impossible for an attacker to correctly guess or brute force the key. Since batch keys provides the strongest protection (with the highest possible key combination), we adopt this key generation method for all the experiments reported in this paper. 

\subsection{Gatekeeper Sign as Digital Signature}
\label{appx:signmethod}
Sign (+/-) of the first hidden state of key $h^{k}_{0}$ can be used to encode a digital signature $S$ such as ASCII code (8 bits as one ASCII character). Note that the maximum capacity of an embedded digital signature depends on the number of hidden units in the protected RNN layer. For instance, in this paper, the model Seq2Seq$_{kt}$ has Gated Recurrent Unit (GRU) layer with 1000 units, so the maximum signature capacity that can be embedded is 1000 bits or 125 ASCII characters. For ownership verification, the embedded digital signature $S$ can be revealed by decoding the learned sign of $h^{k}_{0}$. Table \ref{table:samplesign} shows the embedded digital signature and their respective sign, every 8 bits is decoded into a ASCII character.


\section{Cross Domain Application}
\label{appx:seqmnist}

In addition to NLP domain, we also applied our proposed frameworks on image domain, specifically in the task of sequential image classification. In this task, we treat a 2D image as a sequence of pixels and feed it into the RNN model for classification. This is particularly useful in cases where one cannot obtain the whole image in a single time frame. SeqMNIST \cite{QuocSeqMnist2015} is a variant of MNIST where sequence of image pixels that represent handwritten digit images is classified into 10 digit classes. For trigger sets in image domain, we follow the work by \citet{AdiBackdooring2018} where we select random images from training dataset and shuffle their labels. We chose \citet{QuocSeqMnist2015} as the baseline model and followed the hyperparameters defined in the work which are 100 hidden units in RNN, 128 batch size and Adam \cite{Adam2015} optimizer with default settings.

\subsection{Quantitative and Qualitative Results}
Quantitatively, we achieve similar results as the application in NLP domain. As seen in Table \ref{table:mnistperfmain}, the protected models have similar classification accuracy as the baseline model demonstrating that embedding keys and trigger set doesn't hurt the model learning capacity. Also, we can notice that when a counterfeit key is presented to the protected models, the classification accuracy dropped by 75-80\%. 

Furthermore, we also investigate the qualitative results in sequential image classification task. In Table \ref{table:mnistqualperf}, when a counterfeit key is used, the RNN model gets confused between similar classes, i.e. 5 and 6 for SeqMNIST.

\begin{figure}[t]
\centering
\includegraphics[keepaspectratio=true, scale = 0.35]{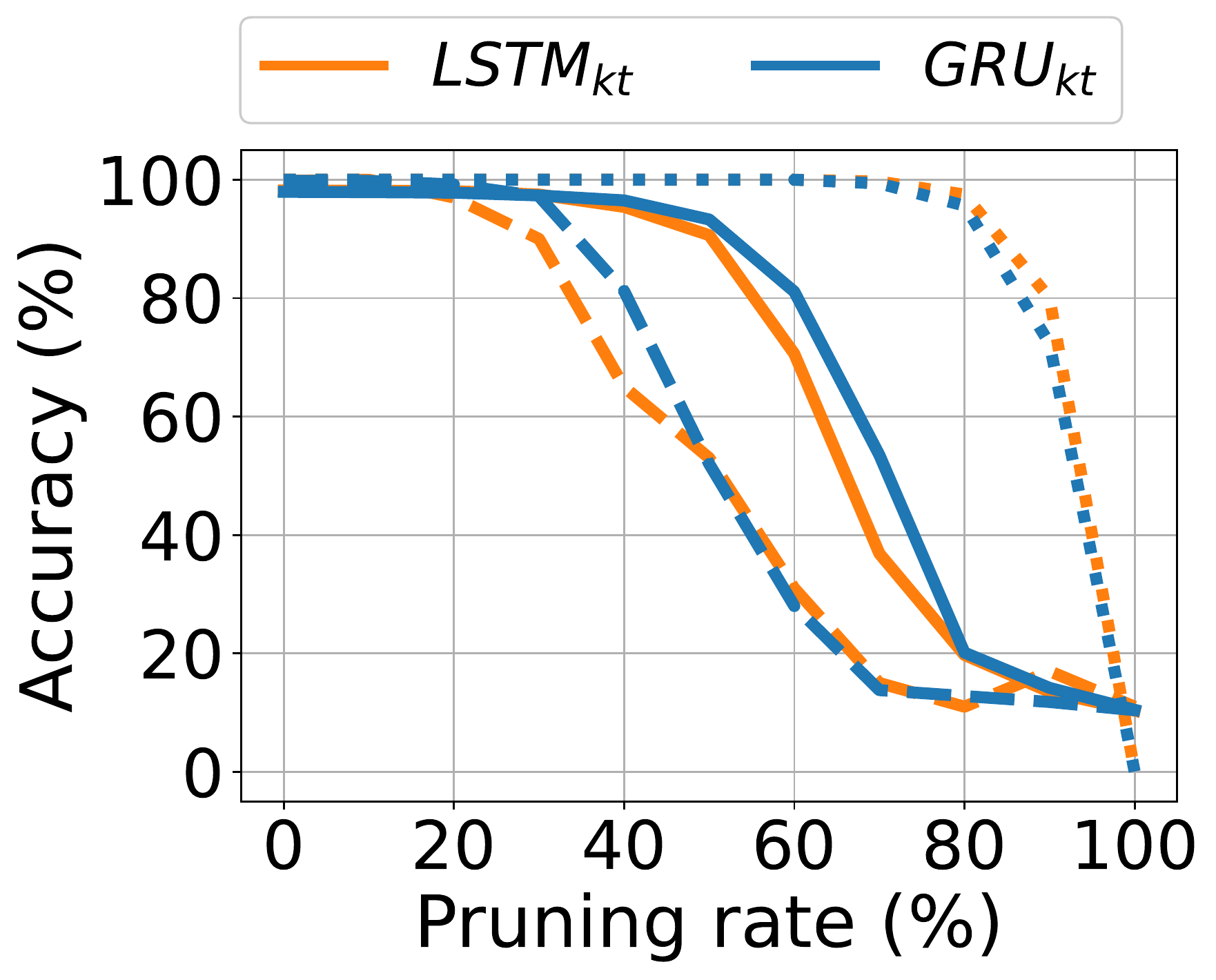}
\caption{Classification accuracy on test set (solid line) and trigger set (dashed line), and digital signature accuracy (dotted line) against different pruning rates for SeqMNIST. Best viewed in colour.}
\label{fig:mnistpruning}
\end{figure}
\begin{figure}[t]
\centering
\includegraphics[keepaspectratio=true, scale = 0.35]{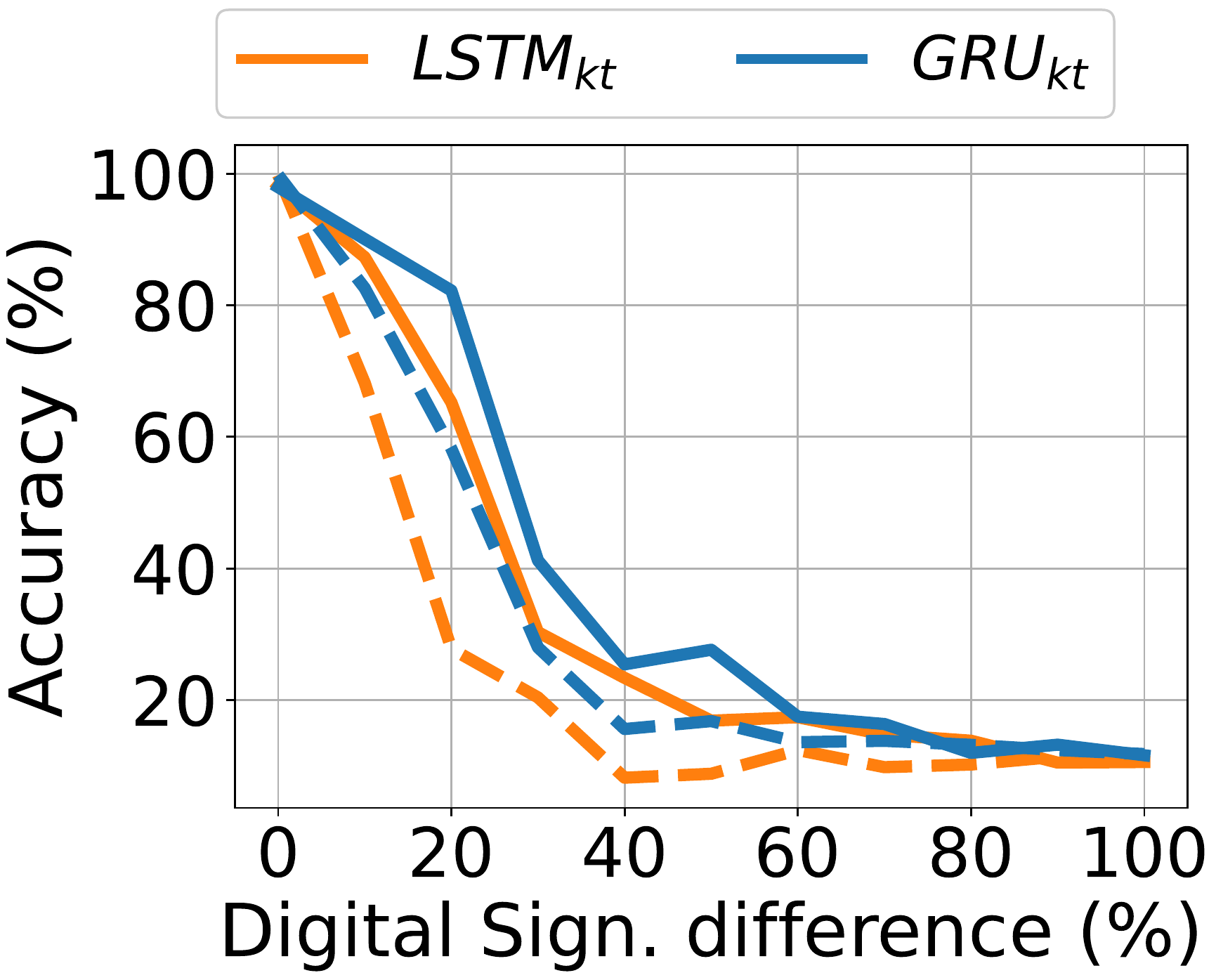}
\caption{Classification accuracy on test set (solid line) and trigger set (dashed line) for SeqMNIST when different percentage (\%) of the digital signature $S$ is being modified/compromised. Best viewed in colour.}
\label{fig:mnistsignflip}
\end{figure}
\begin{figure}[t]
\centering
\includegraphics[keepaspectratio=true, scale = 0.35]{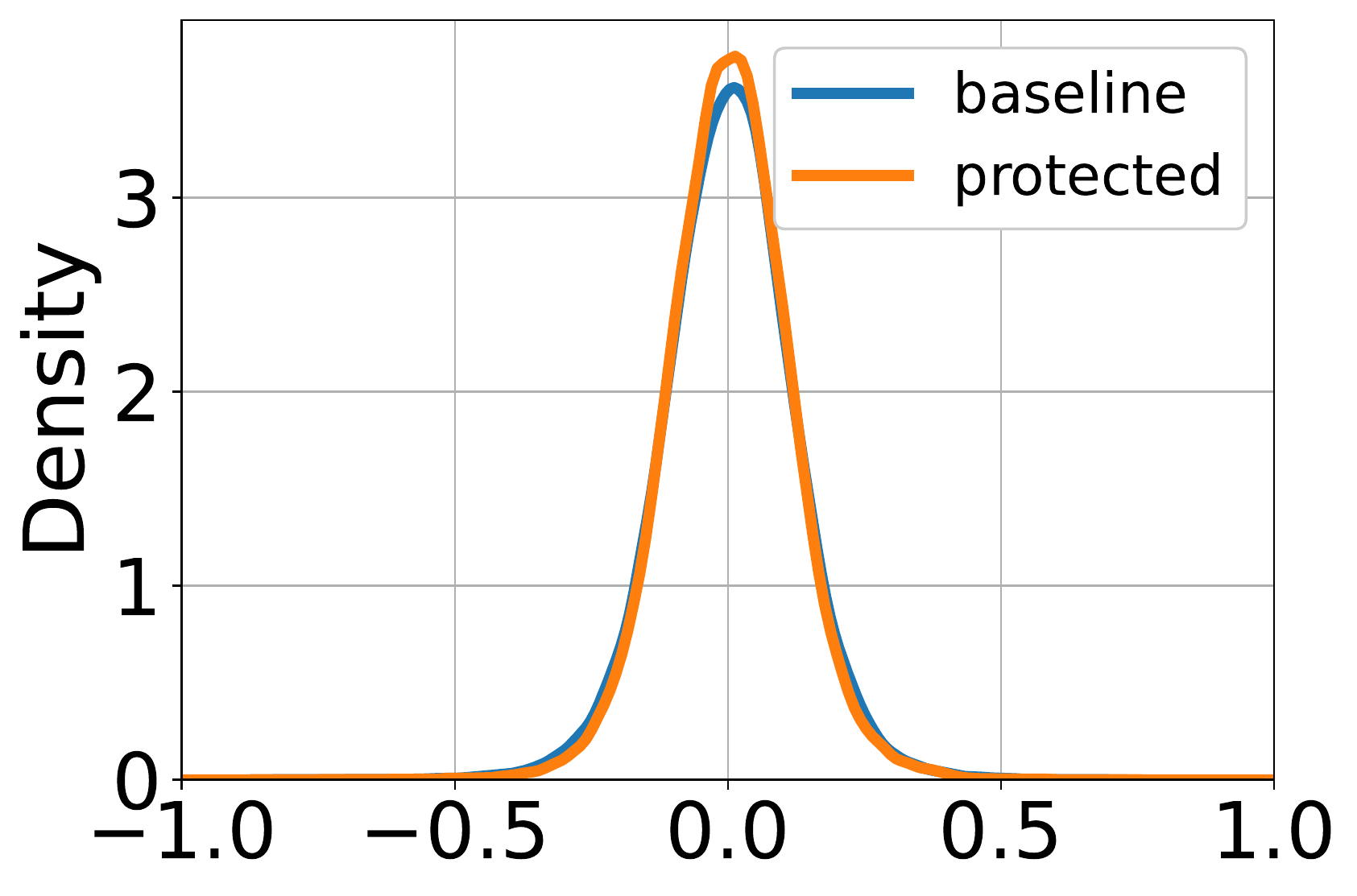}
\caption{Comparison of weight distribution between original and protected model on SeqMNIST. Best viewed in colour.}
\label{fig:mnistweight}
\end{figure}

\subsection{Robustness against Removal Attacks}
\paragraph{Pruning:} We follow the same model pruning strategy in our main paper.
Figure \ref{fig:mnistpruning} shows that for image classification models, even when 40\% of the model parameters are pruned, trigger set accuracy still maintains about 70-80\% accuracy, accuracy on test set drops slightly while digital signature accuracy still maintained near to 100\% accuracy. This proves that model pruning will hurt the model performance before the embedded watermarks can be removed and therefore our proposed work is robust against it.

\paragraph{Fine-tuning:} Same as the main paper, the host model is initialized using trained weights with embedded watermarks, then it is fine-tuned without the presence of the key, trigger set and regularization terms. As seen in Table \ref{table:mnistrem}, digital signature accuracy remains consistently at 100 even after the model is fine-tuned. When the original genuine key is presented to the fine-tuned model, we are able to obtain comparable accuracy to the stolen model.

\paragraph{Overwriting:} We also simulate an overwriting scenario where the attacker has knowledge of how the model is protected and attempts to embed a new key, $\overline{k}$ into the trained model using the same proposed method. In Table \ref{table:mnistrem}, we can observe that digital signature accuracy remains at 100\% consistently after the protected model is overwritten with the new key. When inferencing using the original genuine key, the performance only dropped slightly. Empirically, this confirms that the embedded key and signature cannot be removed by overwriting it with new keys.

\subsection{Resilience against ambiguity attacks}
In the previous section, we simulate a scenario where the key embedding method and digital signature are completely exposed, and an attacker can introduce an ambiguous situation by embedding a new key simultaneously. However, we show that the digital signature cannot be changed easily. As shown in Figure \ref{fig:mnistsignflip}, the model's performance decreases significantly when 40\% of the original signs are modified. In sequential image classification task on SeqMNIST, the model's accuracy dropped from 98.18 $\to$ 23.37 (for the test set in LSTM$_{kt}$), which is merely better than a random guessing model. We can conclude that the signs enforced in this way are persistent against ambiguity attacks and illegal parties will not be able to employ new digital signatures without hurting the protected model's performance.

\subsection{Secrecy}
In digital watermarking for DNN, one of the design goals is secrecy to prevent unauthorized parties from detecting it. In other words, this means that the protected model's weights should not change when compared to a baseline (unprotected) model. Figure \ref{fig:mnistweight} shows the weight distribution of the protected models and baseline model, the weight distribution of the protected RNN layers is identical to the baseline RNN layers. 

\begin{table}[t]
\caption{Robustness of protected RNN model trained on SeqMNIST (in bold) against removal attacks (i.e. fine-tuning and overwriting). All metrics reported herein are the performance with genuine key where acc. = accuracy.}\label{table:mnistrem}
\centering
   \begin{tabular}{l|ccc}
      \toprule
      & Acc. & $T$ acc. & Sign acc. \\ 
      \midrule \midrule
      {\bf LSTM$_{kt}$} & \bf{98.18} & \bf{99.8} & \bf{100} \\ 
      Fine-tuning & 98.28 & 99.6 & 100\\
      Overwriting & 97.52 & 52 & 100\\
      \midrule
      {\bf GRU$_{kt}$} & \bf{97.95} & \bf{99.8} & \bf{100}\\
      Fine-tuning & 98.09 & 99.4 & 100\\
      Overwriting & 97.53 & 78 & 100\\
      \bottomrule
   \end{tabular}
\end{table}

\begin{table}[ht]
\caption{Qualitative results on SeqMNIST. The best-performed model that has both white-box and black-box protections is selected to demonstrate the classification results with genuine and counterfeit keys.}\label{table:mnistqualperf}
\centering
\resizebox{0.9\linewidth}{!}{
  \begin{tabular}{p{0.2\textwidth}|p{0.13\textwidth}|p{0.13\textwidth}|p{0.13\textwidth}}
      \toprule
      Input & Ground Truth & Prediction with genuine key & Prediction with counterfeit key \\
      \midrule \midrule
      \raisebox{-.5\height}{\includegraphics[width=0.15\textwidth]{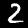}} & 2 & \textcolor{darkgreen}{2 }& \textcolor{red}{7} \\
      \midrule
      \raisebox{-.5\height}{\includegraphics[width=0.15\textwidth]{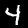}} & 4 & \textcolor{darkgreen}{4} & \textcolor{red}{7} \\
      \midrule
      \raisebox{-.5\height}{\includegraphics[width=0.15\textwidth]{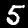}} & 5 & \textcolor{darkgreen}{5} & \textcolor{red}{6} \\
      \midrule
      \raisebox{-.5\height}{\includegraphics[width=0.15\textwidth]{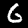}} & 6 & \textcolor{darkgreen}{6} & \textcolor{red}{0} \\
      \midrule
      \raisebox{-.5\height}{\includegraphics[width=0.15\textwidth]{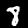}} & 8 & \textcolor{darkgreen}{8} & \textcolor{red}{0} \\
      \bottomrule
  \end{tabular}}
\end{table}


\end{document}